\documentclass[letterpaper, 10 pt, conference]{ieeeconf}

\usepackage{algorithm}
\usepackage[noend]{algpseudocode}
\usepackage{subfig}
\usepackage{siunitx}
\usepackage[most]{tcolorbox}
\usepackage{amsmath}
\usepackage{amsthm}
\usepackage{amsfonts}
\usepackage{url}
\usepackage{xcolor}

\newcommand{\R}{\mathbb{R}}

\newcommand{\Rn}{\mathbb{R}^n}
\newcommand{\Rm}{\mathbb{R}^m}
\newcommand{\Rnm}{\mathbb{R}^{n \times m}}

\newcommand{\Sc}{\mathcal{S}}

\newcommand{\x}{\mathbf{x}}

\newcommand{\uu}{\mathbf{u}}

\newcommand{\xdes}{\mathbf{x}_{\mathrm{des}}}
\newcommand{\y}{\mathbf{y}}

\newcommand{\X}{\mathbf{X}}

\newcommand{\xquery}{\x_*}

\newcommand{\pipex}{\bigg|_{ \xquery}}

\newcommand{\KbarInv}{\mathbf{ \overline{K} \hspace{0.05cm} }^{-1}}

\newcommand{\I}{\mathbf{I}}
\newcommand{\K}{\mathbf{K}}
\newcommand{\Q}{\mathbf{Q}}
\newcommand{\PP}{\mathbf{P}}
\newcommand{\Lm}{\mathbf{\Lambda}}

\newcommand{\fis}{ f_{\mathrm{is}} }
\newcommand{\OmegaExt}{ \Omega_{\mathrm{ext}}}
\newcommand{\OmegaInt}{ \Omega_{\mathrm{int}}}

\newcommand{\pcloud}{\mathcal{P}}
\newcommand{\pbunny}{\pcloud_{\mathrm{bunny}}}
\newcommand{\pchair}{\pcloud_{\mathrm{chair}}}
\newcommand{\pcloudgp}{\mathcal{P}_{\mathrm{gp}}}
\newcommand{\pcloudsgp}{\mathcal{P}_{\mathrm{sgp}}}

\newcommand{\hgp}{ h_{\mathrm{gp}} }
\newcommand{\hsgp}{ h_{\mathrm{sgp}} }

\newcommand{\hb}{ h_{\mathrm{b}} }

\newcommand{\hu}{ h_{\mathrm{u}} }

\newcommand{\hcdot}{h_{\mathrm{(\cdot)}}(\x)}

\newcommand{\unom}{ \uu_{\mathrm{nom}} }
\newcommand{\urect}{ \uu_{\mathrm{rect}} }

\newcommand{\dhsgpdx}{\frac{\partial \hsgp (\x) }{\partial \x}}
\newcommand{\dhcdotdx}{\frac{\partial h_{\mathrm{(\cdot)}}(\x)}{\partial \x}}

\newcommand{\dmudx}{\frac{\partial \mu (\x) }{\partial \x}}
\newcommand{\dvardx}{\frac{\partial \sigma^2 (\x) }{\partial \x}}

\newcommand{\dkdx}{\frac{\partial \mathbf{k} (\x) }{\partial \x}}
\newcommand{\dkmdx}{\frac{\partial \mathbf{k}_M (\x) }{\partial \x}}

\newcommand{\dkmidx}{\frac{\partial \mathbf{k}_{M(i)} (\x) }{\partial \x}}
\newcommand{\ddkddx}{\frac{\partial^2 \mathbf{k} (\x) }{\partial \x^2}}
\newcommand{\ddkiddx}{\frac{\partial^2 \mathbf{k}_{(i)} (\x) }{\partial \x^2}}

\newcommand{\dtdx}{\frac{\partial t (\x) }{\partial \x}}
\newcommand{\ddtddx}{\frac{\partial^2 t (\x) }{\partial \x^2}}

\newcommand{\q}{\mathbf{q}}
\newcommand{\qdot}{\dot{\q}}

\newcommand{\qdotref}{\qdot_{\mathrm{ref}}}

\newcommand{\bigOgp} {\mathcal{O}(N^3)}		
\newcommand{\bigOgpmean} {\mathcal{O}(N)}		
\newcommand{\bigOgpvar} {\mathcal{O}(N^2)}		

\newcommand{\bigOsgp}{\mathcal{O}(M^2N)}		
\newcommand{\bigOsgpmean}{\mathcal{O}(M)}		
\newcommand{\bigOsgpvar}{\mathcal{O}(M^2)}		

\newtheorem{remark}{\textit{Remark}}

\theoremstyle{definition}
\newtheorem{definition}{Definition}

\theoremstyle{problem}
\newtheorem{problem}{Problem}

\theoremstyle{assumption}
\newtheorem{assumption}{Assumption}

\theoremstyle{Example}

\DeclareMathOperator*{\argmin}{arg\,min}

\newcommand\norm[1]{\big\lVert #1 \big\rVert}

\newlength\mylen

\let\oldnl\nl
\newcommand{\nonl}{\renewcommand{\nl}{\let\nl\oldnl}}

\definecolor{green}{rgb}{0,1,0}
\definecolor{lightblue}{rgb}{0,0,0.8} 
\definecolor{red}{rgb}{1,0,0}

\def\fillandplacepagenumber{%
 \par\pagestyle{empty}%
 \vbox to 0pt{\vss}\vfill
 \vbox to 0pt{\baselineskip0pt
   \hbox to\linewidth{\hss}%
   \baselineskip\footskip
   \hbox to\linewidth{%
     \hfil\thepage\hfil}\vss}}

\begin{document}

\IEEEoverridecommandlockouts

\title{Gaussian Process Implicit Surfaces as Control Barrier Functions for Safe Robot Navigation
}


\author{Mouhyemen Khan$^{1}$, Tatsuya Ibuki$^{2}$, Abhijit Chatterjee$^{1}$%
\thanks{$^{1}$School of Electrical and Computer Engineering, Georgia Institute of Technology, Atlanta, GA 30332, USA.}%
\thanks{$^{2}$Department of Electronics and Bioinformatics, School of Science and Technology, Meiji University, Kanagawa 214-8571, Japan.}%
\thanks{Corresponding author: M. Khan, \texttt{mouhyemen.khan@gmail.com}.}%
}



\maketitle

\begin{abstract}
Level set methods underpin modern safety techniques such as control barrier functions (CBFs), while also serving as implicit surface representations for geometric shapes via distance fields. 
Inspired by these two paradigms, we propose a unified framework where the implicit surface itself acts as a CBF.
We leverage Gaussian process (GP) implicit surface (GPIS) to represent the safety boundaries, using \textit{safety samples} which are derived
from sensor measurements to condition the GP. 
The GP posterior mean defines the implicit safety surface (safety belief), while the posterior variance provides a robust safety margin.
Although GPs have favorable properties such as uncertainty estimation and analytical tractability, they scale cubically with data. 
To alleviate this issue, we develop a sparse solution called \textit{sparse Gaussian CBFs}.
To the best of our knowledge, GPIS have not been explicitly used to synthesize CBFs.
We validate the approach on collision avoidance tasks in two settings: a simulated 7-DOF manipulator operating around the Stanford bunny, and a quadrotor navigating in 3D around a physical chair. 
In both cases, Gaussian CBFs (with and without sparsity) enable safe interaction and collision-free execution of trajectories that would otherwise intersect the objects.
\end{abstract}

\maketitle

\section{INTRODUCTION}
Safety-critical robotic systems are rapidly expanding across domains such as
medicine, agriculture, warehouses, disaster response, and defense, where
dependability is crucial since failures can endanger lives and cause major
economic loss. Control barrier functions (CBFs), a level-set approach to
enforcing safety \cite{cbf_theory_Ames2019control}, have been demonstrated on
legged robots \cite{cbf_legged_Grandia2021,cbf_legged_Tend2021}, aerial vehicles
\cite{cbf_cascaded_Khan2019, cbf_racing_drones_Singletary2022}, manipulators \cite{cbf_manipulator_Rauscher2016}, and multi-agent systems \cite{cbf_swarm_consensus_Machida2021}. CBFs guarantee
forward invariance of a prescribed \emph{safe set} using three elements: a
candidate scalar function, a nominal system model, and a nominal control input.
The candidate’s $0$-superlevel set defines the CBF, from which quadratic program
(QP) constraints rectify the nominal input into a safe control. However, most
CBFs are hand-crafted from domain knowledge and heuristics, which is often
infeasible in many real-world applications. \textit{This motivates our goal of achieving safe control for dynamical systems, particularly robots, by constructing CBFs through a data-driven approach.}

Data-driven approaches for constructing CBFs are actively pursued. Expert
demonstrations of state and control trajectories have been used to generate
CBFs \cite{cbf_robust_hybrid_Robey2021}, but relying on such demonstrations is
impractical in unseen environments, and these methods lack hardware validation.
Neural certificates have also been proposed \cite{neural_lyapunov_Gaby2021,neural_convex_Tsukamoto2020}, providing formal
correctness guarantees to learning-based controllers, yet they remain limited to
offline training and simulation. Episodic learning has been applied to update
controllers under safety constraints \cite{cbf_episodic_biped_Csomay2021}, but
the focus was on modeling system uncertainty rather than constructing CBFs.
Support vector machines have been used to characterize CBFs from sensor data
\cite{cbf_supervisedml_Srinivasan2020}, though they require carefully tuned
weights and were demonstrated only in 2D simulation. 
Gaussian processes (GPs) have been used to synthesize CBFs directly from data \cite{cbf_gaussian_Khan2022}. Other studies involved using Bayesian meta-learning for fast adaptation across environments in simulation \cite{cbf_bayesian_onboard_sensor_Hashimoto2023}, and a LiDAR-based GP-CBF for 2D navigation of unicycle robots \cite{cbf_gaussian_lidar_Keyumarsi2024}.
In contrast to \cite{cbf_gaussian_Khan2022,cbf_bayesian_onboard_sensor_Hashimoto2023, cbf_gaussian_lidar_Keyumarsi2024},  we characterize safe implicit surfaces in 3D using GPs, introduce \emph{sparse Gaussian CBFs} to reduce complexity while retaining safety guarantees, and provide hardware results in 3D.

To construct CBFs in 3D from data, we draw on signed distance functions (SDFs) as implicit surface representations. An SDF determines whether a point
$\x \in \Omega$ lies inside or outside the set, and has been shown to be a natural representation for navigation and planning
\cite{is_sdf_map_plan_Oleynikova2016}. SDFs have also been used to construct CBFs with memory \cite{cbf_learning_sdf_Long2021}, though only with 2D sensing and in simulation. In contrast, we employ Gaussian processes (GPs) to form GP implicit surfaces (GPIS) \cite{is_gpis_Williams2006}, which have been studied
for surface representation, navigation, shape estimation, and grasping
\cite{is_gpis_esdf_Wu2021,is_gpis_mapping_Lee2019,is_gpis_shape_grasping_Dragiev2011,is_sparse_gpis_shape_tactile_Gandler2020}.
To the best of our knowledge, GPIS has not been explicitly used to synthesize CBFs.
Motivated by SDFs for surface representation and CBFs for safe control, we
propose a unified framework that directly infers implicit surfaces while
enforcing safety control.

In summary, our \textbf{major contributions} are the following:



\begin{itemize}
\item We introduce a unified approach for implicit surface representation and
safe control, where a GPIS serves as the candidate CBF. To the best of our knowledge, GPIS have not previously been used for synthesizing safe control.

\item We develop \emph{sparse Gaussian CBFs}, which retain the uncertainty
estimation and analytical tractability of Gaussian CBFs while reducing
computational complexity.

\item We validate the approach in both simulation and hardware: a 7-DOF robotic manipulator and a quadrotor platform. To our knowledge, this is the first work to synthesize CBFs entirely from 3D data and deploy them on hardware, enabling collision avoidance with Gaussian and sparse Gaussian CBFs.
\end{itemize}


\section{BACKGROUND PRELIMINARIES}\label{sec:background}
We model safety for a dynamical system using GPs as smooth surfaces, 
combining CBFs and GPIS to design probabilistic implicit 
surfaces. Below, we briefly review CBFs, GPs, and implicit surfaces (see \cite{cbf_Ames2017}, \cite{gp_textbook_Rasmussen2003}, 
\cite{is_textbook_levelset_Osher2004} for details).

\subsection{Control Barrier Function}\label{subsec:cbf}
Consider a general control affine dynamical system,
\begin{align}\label{eq:nonlinear_affine_system}
\dot{\x} &= f(\x) + g(\x) \uu,
\end{align}
where $\x(t) \in \Rn$ is the state and $\uu(t) \in \Rm$ is the control input, $f : \Rn \rightarrow \Rn$ and $g : \Rn \rightarrow \Rnm$ are assumed to be locally Lipschitz continuous. 
Let safety for (\ref{eq:nonlinear_affine_system}) be encoded as the superlevel set $\Sc$ of a smooth function $h : \Rn \rightarrow \R $ as,
\begin{align}\label{eq:safeset}
\mathcal{S} = \{ \x \in \Rn \ | \ h(\x) \geq 0 \}.
\end{align}

\begin{definition}[Control Barrier Function \cite{cbf_Ames2017}]\label{def:cbf}
\textit{The function $h(\x) : \Rn \rightarrow \R$ is defined as a control barrier function (CBF), if there exists an extended class-$\kappa$ function $\alpha$ ($\alpha(0) = 0$ and strictly increasing) such that for any $\x \in \Sc$,
\begin{align}\label{eq:cbf_inequality}
	\sup\limits_{\uu \in \Rm }  L_f h (\x) + L_g h (\x) \uu + \alpha( h (\x)) \geq 0,
\end{align}
}
\end{definition}
\noindent where $L_f h (\x) = \frac{\partial h }{\partial \x}f(\x)$ and $L_g h (\x) = \frac{\partial h }{\partial \x}g(\x)$ are the Lie derivatives of $h(\x)$ along $f(\x)$ and $g(\x)$ respectively.

\subsection{Implicit Surfaces}\label{subsec:imp_surface}
An implicit surface represents the shape of a volumetric object in $n$-dimensional Euclidean space via a function that determines whether a point belongs to the object. Formally, it is defined as the $0$-level set (isosurface) of a real-valued implicit function $\fis : \Rn \rightarrow \R$ for $\x \in \Rn$:
\begin{align}
\fis(\x) \ 
\begin{cases}
> 0, \hspace{0.15cm} &\x \textrm{ outside the surface} \\
= 0, \hspace{0.15cm} &\x \textrm{ on the surface} 	\hspace{2cm} .\\
< 0, \hspace{0.15cm} &\x \textrm{ inside the surface}
\end{cases}
\end{align}

The $0$-level set $\fis^{-1}(0)$ of an implicit function $\fis$ defines a surface $\Sc \subset \Rn$, which in our case represents the safety boundary. Conversely, for any surface $\Sc$ in $\Rn$, there exists a function $\fis : \Rn \to \R$ such that $\fis^{-1}(0) = \Sc$ (Prop.~2 in \cite{is_textbook_Do2016}). Hence, given samples on $\Sc$, one can construct $\fis$. We approximate the implicit surface of a volumetric object in $d=3$ ($d \leq n$) using a GP regressor, referred to in the literature as a Gaussian Process Implicit Surface (GPIS).

\subsection{Gaussian Process Regression}\label{subsec:gp}
GP regression is a well established nonparametric approach which relies on kernels for solving non-linear regression tasks. Kernels provide a notion of similarity between pairs of input points, $\x_i, \x_j \in \Rn$. A popular kernel is the squared exponential (SE) kernel given by,
\begin{align}\label{eq:gaussian_kernel}
\hspace{-0.3cm}
k(\x_i,\x_j) 
\hspace{-0.05cm}
=
\sigma_f^2 \exp 
\hspace{-0.1cm} 
\bigg( \hspace{-0.15cm} 
- 
\frac{ ( \x_i - \x_j )^{\top} 
\mathbf{L}^{-2} 
( \x_i - \x_j) } {2}  
\hspace{-0.1cm}
\bigg) 
\hspace{-0.1cm} 
+ 
\hspace{-0.05cm} 
\delta_{ij} \sigma_y^2 ,
\hspace{-0.1cm}
\end{align}
where $\delta_{ij} = 1$ if $i=j$ and $0$ otherwise, $\mathbf{l} \in \Rn$ is the characteristic length scale, with $\mathbf{L} = \mathrm{diag}(\mathbf{l}) \in \R^{n \times n}$. 
The signal scale and observation noise are $\sigma_f^2$ and $\sigma_y^2$ respectively. Together, these parameters constitute the SE kernel's hyperparameters, $\Theta = \{ \mathbf{L}, \sigma_f^2, \sigma_y^2 \}$. These hyperparameters for a dataset can be optimized by maximizing the log marginal likelihood using quasi-Newton methods \cite{gp_textbook_Rasmussen2003}.

We are interested in constructing an implicit surface, which will later be our safety function of interest, for which we will carefully design scalar targets $y$ in Section \ref{subsec:data_processing}.
Given a set of $N$ data points, with input vectors $\x \in \Rn$, and scalar targets $y \in \R$, we compose the dataset $\mathcal{D}_N = \{ \X_N, \y_N \}$, where $\X_N = \{\x_i\}_{i=1}^{N}$ and $\y_N = \{y_i\}_{i=1}^N$. GPs can compute the posterior mean and variance for an arbitrary deterministic query point $ \xquery \in \Rn$, by conditioning on previous measurements. The posterior mean $\mu \in \R$ and variance $\sigma^2 \in \R$ are \cite{gp_textbook_Rasmussen2003},
\begin{align}
\mu( \xquery) &= \mathbf{k}( \xquery)^{\top} \ \KbarInv \y_N, \label{eq:gp_mean} \\
\sigma^2( \xquery) &= k ( \xquery,  \xquery) - \mathbf{k}( \xquery)^{\top} \ \KbarInv \mathbf{k}( \xquery) \label{eq:gp_var},
\end{align}
where $\mathbf{k}(\xquery) = \big[k(\x_1,  \xquery), \ldots , k(\x_N,  \xquery) \big]^{\top} \in \mathbb{R}^{N}$ is the covariance vector between $\X_N$ and $ \xquery$, $\mathbf{\overline{K}} \in \R^{N \times N}$, with entries $[ \overline{k} ]_{(i,j)} = k(\x_i, \x_j), \ i, j \in \{1, \ldots, N\}$, is the covariance matrix between pairs of input points in $\X_N$, and $k( \xquery,  \xquery) \in \R$ is the prior covariance. 

\section{PROBLEM STATEMENT}\label{sec:problem}
\begin{figure}[!t]
\centering
\includegraphics[width=1\linewidth]{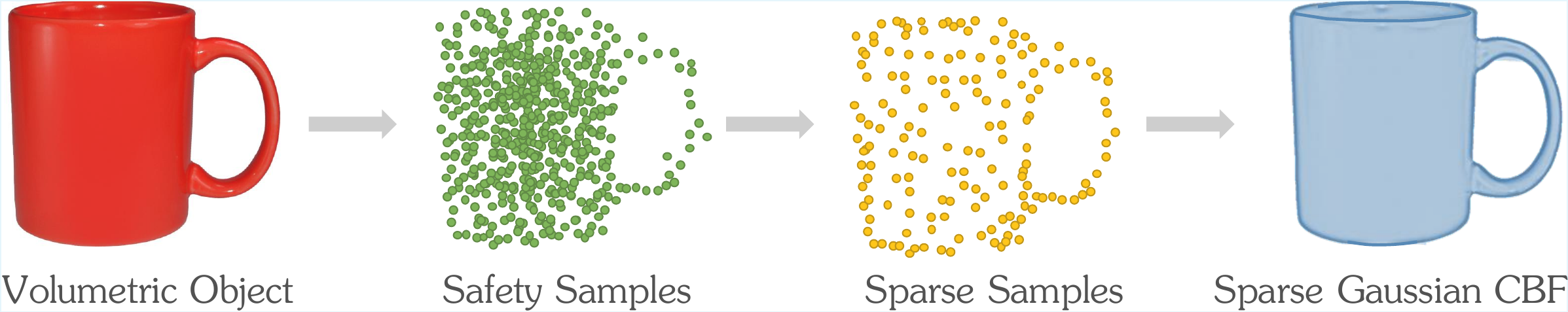}
\vspace{-0.5cm}
\caption{The safety surface of a volumetric object of interest is modeled as a sparse Gaussian CBF using safety samples.}
\label{fig:sgcbf_pipeline}
\vspace{-0.5cm}
\end{figure}

Given a dynamical system \eqref{eq:nonlinear_affine_system} and a $3$D object, our goal is to recover the object’s surface using GPs and perform collision avoidance through safe control. The surface is learned from on- and off-surface points derived from sensor data, which we term safety samples (Fig.~\ref{fig:sgcbf_pipeline}), and these samples define the Gaussian CBF.

\begin{assumption}\label{assump:sensor_safety_samples}
We assume access to sensor data providing surface points and normals of the
object.
\end{assumption}

This assumption is reasonable, since many sensing modalities (e.g., laser, haptic, or
vision-based) provide point clouds and surface normals
\cite{is_gpis_shape_grasping_Dragiev2011}. 

\begin{remark}\label{remark:dimension_difference}
Safety surface is modeled in $d=3$ (Euclidean space) while the
system state $\x \in \Rn$ with $d \leq n$, the $(n-d)$ free dimensions can be
set to zero during GP training.
\end{remark}
\vspace{-.6cm}
\begin{align}
\tcboxmath{
\hspace{-0.4cm}
\underbrace{\fis (\x) }_{\text{Implicit Surface}} 
\hspace{-0.1cm}
= 
\hspace{-0.1cm}
\underbrace{ \hsgp (\x)}_{\text{Sparse Gaussian CBF}} 
\hspace{-0.3cm}
}
:= 
\hspace{-0.1cm}
\underbrace{\hb(\x)}_{\text{Safety Belief}}
\hspace{-0.1cm}
+ 
\hspace{-0.1cm}
\underbrace{\hu(\x)}_{\text{Safety Margin}}
\hspace{-0.1cm}
.
\hspace{-0.15cm}
\end{align}

The sparse Gaussian CBF provides an implicit surface approximation of a volumetric object. Because this surface is data-driven, it is necessary to include a safety margin in the safety function estimate, especially when exploiting sparsity, where uncertainty must be addressed. The safety belief denotes the best estimate of system safety (or the implicit surface), while the safety margin captures uncertainty in the estimate, increasing in regions with limited data. 

\begin{problem} 
Given system \eqref{eq:nonlinear_affine_system} and online measurements of the system state $ \xquery$, synthesize $\hsgp(\x)$ to model an implicit surface with a safety belief and associated safety uncertainty, conditioned on past safety samples and observations for a volumetric object in the dataset, $\mathcal{D}_N = \{ \X_N, \y_N \}$, where $\X_N \hspace{-0.1cm}=\hspace{-0.1cm} \{\x_i\}_{i=1}^{N}$ and $\y_N \hspace{-0.1cm}=\hspace{-0.1cm} \{y_i\}_{i=1}^N$, such that system (\ref{eq:nonlinear_affine_system}) is safe.
\end{problem}

Given a nominal controller, the system may not be safe under its influence. Hence, we want to modify the nominal input using the synthesized $\hsgp(\x)$ to achieve safety.

\begin{assumption}\label{assump:nominal_control}
We assume a nominal control input $\unom$ exists that drives system \eqref{eq:nonlinear_affine_system} state $\x$ to a desired state $\xdes$.
\end{assumption}

\begin{problem}
Given system \eqref{eq:nonlinear_affine_system}, the synthesized $\hsgp(\x)$ with safe set $\Sc$, and nominal control input $\unom \in \R^m$, rectify the control input $\urect \in \R^m$ ensuring system (\ref{eq:nonlinear_affine_system}) is safe.
\end{problem}


However, designing $\urect$ with a nonparametric CBF is difficult, since Lie derivatives are ill-posed without assumptions. We address this by using a kernel representation to encode both the safety belief and its margin. 

\section{PROPOSED METHODOLOGY}\label{sec:sparse_gcbf}
Here, we present our approach to approximate the GPIS $\fis$ with a sparse Gaussian CBF $\hsgp(\x)$. GPs are preferable to models such as neural networks, polynomial chaos, or radial basis functions because their Bayesian nonparametric formulation provides flexible priors and closed-form probabilistic posterior estimates. Moreover, GPs naturally yield uncertainty through posterior variance, whereas neural networks require careful design to capture uncertainty \cite{nn_survey_Gawlikowski2021}.

\subsection{Data Processing}\label{subsec:data_processing}
Given $N_s$ sensor measurements in an $n$-dimensional Euclidean space,
$\mathcal{P} \in \R^{N_s \times n} \times \R^{N_s \times n}$ denotes the set of point-cloud data and corresponding surface normals. 
The point-cloud gives the on-surface points, while off-surface points , both external and internal, are generated using the surface normals; collectively called \textit{safety samples},
\begin{align*}
\OmegaExt &:= \{ \ \x_{+} \ \ \hspace{0.02cm}  | \ \ \fis(\x_{+}) > 0\} \\
\Omega_0  &:= \{ \ \x_{0}  \ \ \hspace{0.08cm} | \ \ \fis(\x_{0}) = 0\} \hspace{0.5cm} .\\
\OmegaInt &:= \{ \ \x_{-} \ \ | \ \ \fis(\x_{-}) < 0\}
\end{align*}
Since only the point-cloud coordinates and surface normals in $\mathcal{P}$ are available, we construct the sets $\Omega_0$, $\OmegaExt$, and $\OmegaInt$.

\begin{itemize}
    \item $\Omega_0$ is formed by taking $N_0$ points from the point cloud 
    $\mathcal{P}$ ($N_0 \leq N_s$), each labeled with $0$, i.e., 
    $\y_{N_0} = \{0\}_{i=1}^{N_0}$.
    \item $\OmegaExt$ is formed by taking $N_+$ synthetic points 
    ($N_+ < N_0$) along the surface normals with a fixed offset.  
    \item $\OmegaInt$ is formed by creating $N_-$ points 
    ($N_- \leq N_+$) in the opposite normal direction.
    \item Targets are $+1$ for $\OmegaExt$ and $-1$ for $\OmegaInt$, creating the labels,  
    $\y_{N_+} = \{+1\}_{i=1}^{N_+}$ and 
    $\y_{N_-} = \{-1\}_{i=1}^{N_-}$.
    \item The GP training set is  
    $\X_N \hspace{-0.2cm}=\hspace{-0.1cm} [\x_0^{\top}, \x_{+}^{\top}, \x_{-}^{\top}]^{\top} \hspace{-0.2cm}\in \R^{N \times n}$  
    with labels  
    $\y_N \hspace{-0.1cm}=\hspace{-0.1cm} [\y_{N_0}, \y_{N_+}, \y_{N_-}]^{\top} \hspace{-0.2cm}$,  
    where $N \hspace{-0.1cm}=\hspace{-0.1cm} N_0 + N_+ + N_-$.  
\end{itemize}
\subsection{Sparse Gaussian Control Barrier Function}\label{subsec:sparse_gcbf}
We use a GP prior on the desired safety function, $\hsgp \sim \mathcal{GP}(0, k(\x, \x'))$ where the inputs and targets to the GP are $\X_N$ and $y_N$ as discussed above. By using the GP prior, we fully specify the candidate safety function, unlike \cite{cbf_gaussian_variance_Khan2021} where only the posterior variance was modeled by GP. Additionally, we  also present a sparse approximation to our approach.

\begin{assumption}\label{assump:initialize_safeset}
The safe set is nonempty with at least one data point, the initial state $\x(0)$ and associated safety value $\hsgp(\x(0)) \in \R_{\geq 0}$, to synthesize $\hsgp$.
\end{assumption}

We assume that the system begins in an initial compact safe set. Safety for $\hsgp$ is encoded as,
\begin{align}
\Sc = \{ \x \in \Rn \ | \ \hsgp(\x) \geq 0 \}, \label{eq:sgcbf_safeset} \\
\partial \Sc = \{ \x \in \Rn \ | \ \hsgp(\x) = 0 \}. \label{eq:sgcbf_safeset_boundary}
\end{align}

Despite GPs being very powerful regressors, as the dataset grows larger, they become computationally intractable. 
GP prediction complexity has a cost of $\bigOgp$. Even if one stores the covariance matrix to save costs, the complexity per test case is $\bigOgpmean$ for predictive mean and $\bigOgpvar$ for predictive variance. 
Hence, many sparse approximations of GPs have been developed to reduce the complexity cost while retaining accuracy \cite{gp_power_ep_Minka2001, gp_sparse_fitc_Snelson2006, gp_sparse_vfe_Titsias2009}. 
We focus on the variant called \textit{Sparse Pseudo-Input Gaussian Process} (SPGP) \cite{gp_sparse_fitc_Snelson2006}. Here, we simply present its mean and variance:
\begin{align}
\mu(\xquery) &= \mathbf{k}^{\top}_{M} (\xquery) \ \Q_M^{-1} \ \K_{MN} \big( \Lm_N + \sigma_y^2 \I_N \big)^{-1} \y_N  \label{eq:sgp_mean}\\
\sigma^2(\xquery) &= k (\xquery, \xquery) - \mathbf{k}^{\top}_{M} (\xquery) \PP_M  \mathbf{k}_{M} (\xquery) \label{eq:sgp_var},
\end{align}
where $\hspace{-0.15cm} \mathbf{k}_M(\xquery) \hspace{-0.3cm} = \hspace{-0.3cm} \big[k(\overline{\x}_1,  \xquery), \ldots , k(\overline{\x}_M,  \xquery) \big]^{\top} \hspace{-0.5cm} \in \mathbb{R}^{M}$ is the covariance between $\overline{\X}_M$ and $ \xquery$, 
$\Q_M =  \K_{NM}^{\top} \big( \Lm_N + \sigma_y^2 \I_N  \big)^{ -1} \K_{NM} + \K_M \in \R^{M \times M}$,  
$\Lm_N = \text{diag} \big[ \K_N \ - \  \K_{NM} \ \K_M^{ \ -1} \ \K_{MN} \big] \in \R^{N \times N}$ is a diagonal matrix, 
$\K_M \in \R^{M \times M}$ is the covariance between pairs of pseudo-inputs $\overline{\X}_M$, $\PP_M = \K_M^{ \ -1} - \Q_M^{ \ -1}$, and $\big[ \K_{NM} \big]_{(i,j)} = k(\x_i, \overline{\x}_j), \ i \in \{1, ... , N\}, \  j \in \{1, ... , M\}$ is the covariance matrix between $\X_N$ and $\overline{\X}_M$. 
Computation cost for $\Q_M$ is dominated by the inversion operation which is $\bigOsgp$ \cite{gp_sparse_fitc_Snelson2006}. By precomputing the inverse, the cost per test case is $\bigOsgpmean$ and $\bigOsgpvar$ for predictive mean and variance respectively. We can jointly optimize for the kernel hyperparameters, $\Theta$, and pseudo-inputs, $\overline{\X}_M$, by maximizing the log marginal likelihood using quasi-Newton gradient methods:
\begin{align}\label{eq:sparse_log_mll}
\hspace{-0.2cm}
\log p(y &| \X, \overline{\X}) 
\hspace{-.1cm}
= 
\hspace{-.1cm}
\frac{
\hspace{-.1cm}
- N \log(2\pi) 
+ 
\log | \overline{\K}_N | 
+ 
\y_N \overline{\K}_N^{ \ -1} \y_N}{2}
\hspace{-.2cm}
\end{align}

We propose the following sparse Gaussian CBF $\hsgp(\x)$ which models both the safety belief and associated margin using the sparse predictive mean \eqref{eq:sgp_mean} and variance \eqref{eq:sgp_var},
\begin{align}\label{eq:sparse_gcbf}
\hsgp(\x) &:= \mu(\x) + \sigma^2(\x) \notag\\
& = \underbrace{\mathbf{k}_M^{\top}(\x) \overline{\Q}_{MN} \y_N}_{\textrm{safety belief}}  
+ 
\underbrace{k (\x, \x) - \mathbf{k}^{\top}_{M} (\x) \mathbf{P}_{M} \mathbf{k}_{M} (\x)}_{\textrm{safety margin}},
\end{align}
where $\overline{\Q}_{MN} = \Q_M^{-1} \ \K_{MN} \big( \Lm_N + \sigma_y^2 \I_N \big)^{-1} \in \R^{M \times N}$. 
\textit{As can be seen from the formulation above, we fully realize the safety function using GPs, as well as exploit sparsity, to model any arbitrary volumetric object in an $n$-dimensional Euclidean space while addressing safety.} The admissible control space for the sparse Gaussian CBF ensures that the system \eqref{eq:nonlinear_affine_system} remains forward invariant in the safe set characterized by $\hsgp$ (Theorem 1 in \cite{cbf_Ames2017}). 

\begin{remark}
Sparse Gaussian CBFs offer several advantages for modeling implicit surfaces 
with safety considerations. First, sensor data directly informs the safety 
function, enabling data-driven surfaces rather than hand-crafted candidates. 
Second, GPs yield posterior variance, providing principled uncertainty 
estimates for predictions. Third, sparsity ensures real-time feasibility 
when handling large datasets.
\end{remark}

\subsection{Lie Derivatives of Sparse Gaussian CBF}\label{subsec:sparse_lie_derivatives}
To achieve safe control using sparse Gaussian CBFs, we require the necessary Lie derivatives for ensuring forward invariance in the safe set. A key advantage of using GPs is the use of kernel representations. This makes it easy to compute the partial derivatives in closed-form analytically for the safety belief and margin using \eqref{eq:sgp_mean}, \eqref{eq:sgp_var}. The partial derivative of \eqref{eq:sparse_gcbf} with respect to $\x$ at a query point $\xquery$ is,
\begin{align}
\hspace{-.3cm}
\dhsgpdx \pipex \hspace{-0.3cm}
\hspace{-.1cm}
&= 
\dmudx \pipex + \dvardx \pipex \notag \\
\hspace{-.1cm}
&= \bigg( \y_N^{\top} \overline{\Q}_{NM} 
- 
2 \mathbf{k}_M^{\top}( \xquery) \mathbf{P}_M \bigg) \dkmdx \pipex
\hspace{-.3cm}
,
\label{eq:dhsgpdx}
\end{align}
where $\overline{\Q}_{MN}$ and $\mathbf{P}_M$ are defined in \eqref{eq:sparse_gcbf}. The derivative of the SE kernel \eqref{eq:gaussian_kernel} in (\ref{eq:dhsgpdx}) is given by,
\begin{align}\label{eq:kernel_deriv}
\dkmidx \pipex &= (\x_{(i)} -  \xquery)^{\top} \ k(\x_{(i)} ,  \xquery) \mathbf{L}^{-2} ,
\end{align}
where $\mathbf{k}_{(i)}$ is the $i^{\mathrm{th}}$ element of $\mathbf{k}_M(\x)$, and (\ref{eq:kernel_deriv}) is the $i^{\text{th}}$ row of $\dkmdx \in \mathbb{R}^{M \times n}$. Now, we can compute the Lie derivatives of $\hsgp(\x)$ by taking its time derivative as follows,
\begin{align}
\dot{h}_{\mathrm{sgp}}(\x) 
	&= \dhsgpdx f(\x) + \dhsgpdx g(\x) \uu		\notag 	\\
	&= L_f \hsgp (\x) + L_g \hsgp (\x) \uu, \label{eq:sgcbf_lie}
\end{align}
\noindent where (\ref{eq:dhsgpdx}) is used to get $L_f \hsgp (\x)$ and $L_g \hsgp (\x)$.

\subsection{Safe Control using Sparse Gaussian CBF}\label{subsec:sparse_safe_control}

Given $\unom$ and $\xdes$ outside the safe set $\Sc$, the system would exit $\Sc$ and violate safety. To ensure forward invariance, $\unom$ is rectified by solving an online QP with constraints from the Lie
derivatives in \eqref{eq:sgcbf_lie} \cite{cbf_Ames2017}:

\begin{algorithm}
  Sparse Gaussian CBF QP: \textit{Input modification}
	\begin{align}\label{eq:sgcbf-qp_degree1}
		 \urect &= \argmin_{ \uu \in \Rm} \frac{1}{2}  \norm{ \uu - \unom }^2 \ \ \textrm{subject to} \\[-0.7ex] 
			 & \ \  L_f \hsgp (\x) + L_g \hsgp (\x) \uu + \alpha( \hsgp (\x)) \geq 0, \notag
	\end{align}
\end{algorithm}

\noindent where $\uu_\mathrm{rect}$ is the rectified control input. 
This guarantees forward invariance of the system within the safe set $\Sc$. The procedure for generating the safe control input using the synthesized CBF is summarized in Algorithm~\ref{algo:sgcbf_algorithm}.

\begin{algorithm}
\caption{Sparse Gaussian CBF Synthesis \& Safe Control}\label{algo:sgcbf_algorithm}

\textbf{Input:} 

\hspace*{\algorithmicindent}  \textsc{System} \eqref{eq:nonlinear_affine_system} \& \textsc{Nominal Input} $\unom$

\hspace*{\algorithmicindent}  \textsc{GP Prior} $\hsgp(\x) \sim \mathcal{GP}(0, k(\x,\x'))$


\hspace*{\algorithmicindent}  \textsc{Dataset} $\mathcal{D}_N$ \& number of pseudo-points $M$



\begin{algorithmic}[1]
\Procedure{SafeControl}{}
      	\State \textsc{Initialize} $\mathcal{D}_M = \{ \overline{\X}_M, \overline{\y}_M \}$ randomly from $\mathcal{D}_N$
      	\State \textsc{Optimize} $\hsgp(\X_N, \y_N ; \overline{\X}_M, \overline{\y}_M )$ using \eqref{eq:sparse_log_mll}
      	\State \textsc{Synthesize} $\hsgp(\X_N, \y_N)$ using \eqref{eq:sparse_gcbf}
      	\State \textsc{Compute} $\dhsgpdx$ using \eqref{eq:dhsgpdx} \& \eqref{eq:kernel_deriv}
      	\State \textsc{Setup QP} constraint using \eqref{eq:nonlinear_affine_system} \& \eqref{eq:sgcbf_lie}
      	\State \textsc{Rectify} $\unom$ using \eqref{eq:sgcbf-qp_degree1}
      	
\hspace{-0.8cm} \textbf{return} $\urect$
\EndProcedure
\end{algorithmic}
\end{algorithm}

A subset of $\mathcal{D}_N$ is first sampled to initialize the pseudo-dataset $\mathcal{D}_M$. Hyperparameters and pseudo-inputs are optimized in step~3, followed by the synthesis of the sparse Gaussian CBF in step~4, which defines the safety surface for control. The corresponding derivatives are computed in steps~5--6 to construct the QP constraint, and the nominal control input is rectified in step~7.

\begin{remark}
Owing to its nonparametric nature, the sparse Gaussian CBF can be treated as a 
black-box model for control synthesis, allowing Algorithm~\ref{algo:sgcbf_algorithm} 
to operate without requiring an explicit reformulation of the Lie derivatives. 
Unlike traditional CBFs, where altering the function changes the derivative form, 
Gaussian CBFs preserve the same structure, with their characterization depending 
only on the data and the dynamical system.
\end{remark}

\section{TEST CASE I : MANIPULATOR SIMULATION}\label{sec:test_case1_manip_bunny}
We apply Gaussian CBFs, with and without sparsity, to control a 7-DOF robot manipulator in simulation. Such manipulators are widely used in warehouses, manufacturing, and medicine, where safe manipulation is essential.

\begin{figure}[!b]
\centering
\vspace{-0.25cm}
\includegraphics[width=1\linewidth]{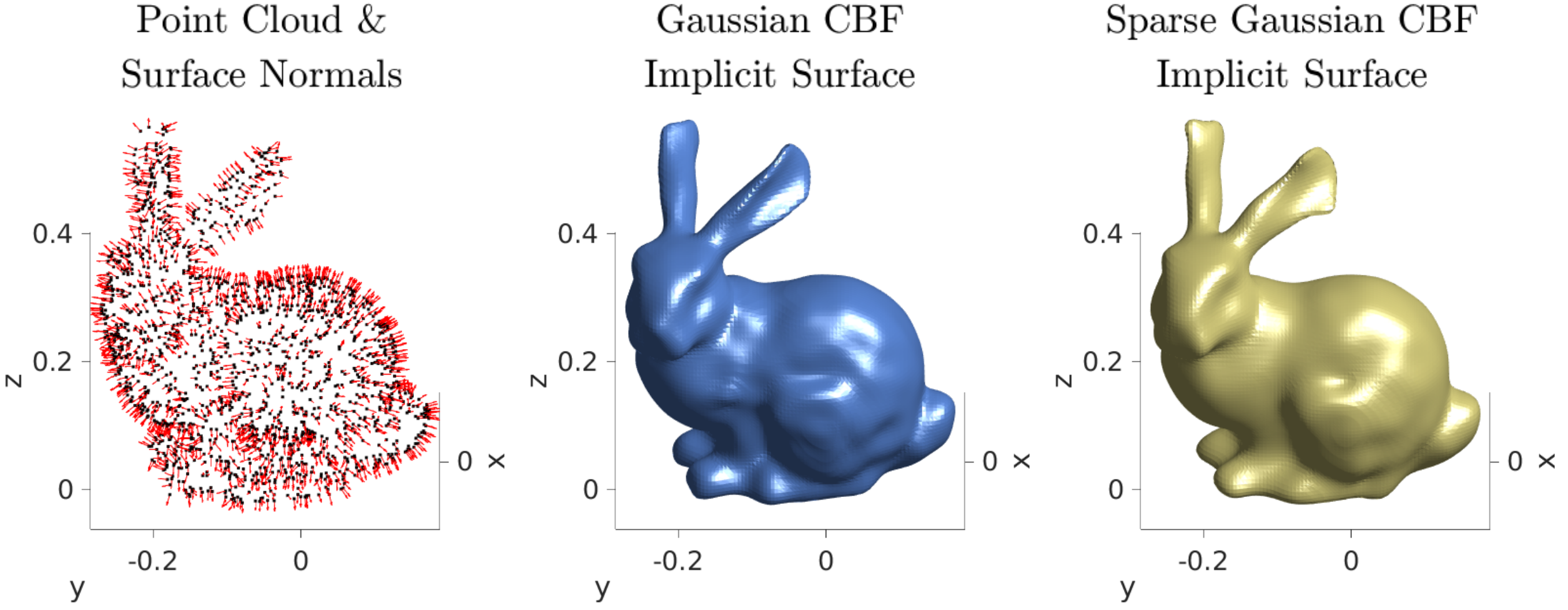}
\caption{The Stanford bunny defines the safe set boundary via Gaussian CBFs: (left) point cloud with surface normals, (middle) Gaussian CBF, 
(right) sparse Gaussian CBF.}
\label{fig:gpis_stanford_bunny}
\end{figure}

\subsection{Stanford Bunny Implicit Surface Modeling}\label{subsec:gpis_bunny}

We model the Stanford bunny surface as the safe set boundary using both full and sparse GPs. The bunny, a standard 3D graphics test model, provides a complex non-convex geometry for safety evaluation. From $34{,}817$ data points, we down-sampled to $3{,}500$ and scaled the model to a $0.45 \si{\m} \times 0.45 \si{\m} \times 0.55 \si{\m}$ cuboid.

\subsubsection{Offline Training} The safety function is synthesized as the implicit surface of the bunny, with the Gaussian CBF formulation given in Section~\ref{subsec:safe_control_manipulator}. Training was performed with the \texttt{gpml} toolbox~\cite{gp_toolbox_Rasmussen2010} 
on an Intel i7-9800X CPU (16 GB RAM, 4.4 \si{\GHz}). For the Gaussian CBF, $2{,}178$ surface points and normals were sampled, giving an average training time of $14.15 \si{\s}$. The sparse Gaussian CBF used one-fifth of these as pseudo-points, reducing training time to $9.46 \si{\s}$. Online training is described in Section~\ref{subsec:proximal_bunny_sim}. Figure \ref{fig:gpis_stanford_bunny} shows the $0$-isosurfaces: the Gaussian CBF captures fine facial contours, while the sparse variant preserves shape with less detail, showing the accuracy–speed trade-off.

\subsubsection{Evaluation Metric} We quantify the modeling performance using the Chamfer distance, a symmetric metric that measures the distance between two point clouds $\pcloud_1$ and $\pcloud_2$ sampled from the surfaces.
\begin{align}\label{eq:chamfer_distance}
\hspace{-0.3cm}
d_{\mathrm{ch}}(\pcloud_1, \pcloud_2) 
\hspace{-0.1cm}
=
\hspace{-0.2cm} 
\sum_{x \in \pcloud_1} \hspace{-0.2cm}\argmin_{\y \in \pcloud_2} \norm{ \x - \y}  
\hspace{-0.1cm}+\hspace{-0.2cm}
\sum_{\y \in \pcloud_2} \hspace{-0.2cm}\argmin_{\x \in \pcloud_1} \norm{\y - \x}
\hspace{-0.3cm}\end{align}

We compute the Chamfer distance between the bunny point cloud $\pbunny$ and 
the $0$-isosurfaces of $\hgp$ and $\hsgp$, obtaining 
$d_{\mathrm{ch}}(\pbunny,\pcloudgp)=0.011$ and 
$d_{\mathrm{ch}}(\pbunny,\pcloudsgp)=0.042$. These values indicate good surface approximation, with the GP capturing finer detail than the sparse model.

\subsection{Robot Manipulator Kinematics}
\label{subsec:robot_manipulator_kinematics}

We control a 7-DOF manipulator’s end-effector around complex volumetric objects (e.g., the bunny) using kinematic control with direct joint-velocity inputs. The dynamics are  
\begin{align}\label{eq:rom_kinematics}
    \qdot = \uu, \quad 
    \unom = \qdotref,
\end{align}  
where $\q \in \R^7$ are the joint positions and $\uu \in \R^7$ the control inputs. Reference velocities are obtained by interpolating the end-effector’s homogeneous transformation matrix between initial and desired poses. We simulate the Kinova Gen3 manipulator in MATLAB's Robotic Systems Toolbox, creating trajectories that pass through the bunny. The synthesized CBFs and reference trajectories are shown in 
Figure~\ref{fig:bunny_robot_reference}.

\subsection{Safe Kinematic Control Synthesis for Manipulator}\label{subsec:safe_control_manipulator}
We rectify the nominal joint-velocity control to enforce safety while the manipulator tracks its reference trajectory around the bunny without 
collision. Safety is ensured using Gaussian CBFs (with and without sparsity) trained to represent the bunny’s implicit surface. Each candidate CBF is  
\begin{align}\label{eq:candidate_gcbf_bunny}
    \hcdot &= \mu(\x) + 4\sigma^2(\x), \\
    \dot{h}_{(\cdot)}(\x) 
    &=
    \dhcdotdx \frac{\partial \x}{\partial \q}\dot{\q} 
    = \dhcdotdx \, \mathbf{J}(\q)\dot{\q} \ ,
\end{align}  
where $\hcdot$'s $0$-isosurface is the bunny, $\mathbf{J}(\q):\R^7 
\to \R^{3 \times 7}$ is the manipulator Jacobian, and $\dhcdotdx$ is computed 
from \eqref{eq:dhsgpdx} (sparse GP) or \eqref{eq:gp_mean}–\eqref{eq:gp_var} 
(regular GP). The QP safety constraint is, 
\begin{align}
\underbrace{\dhcdotdx \ \mathbf{J}(\q)}_{L_g \hcdot } \ \underbrace{\qdot}_{\uu} &\geq -k_0 \hcdot. \label{eq:manipulator-lie_derivative_qp_degree1}
\end{align}

With the Lie derivatives, and the decision variable, $\uu = \qdot$, appearing linearly in the inequality \eqref{eq:manipulator-lie_derivative_qp_degree1}, we can rectify the nominal control, $\unom = \qdotref$, using the QP in \eqref{eq:sgcbf-qp_degree1}.

\subsection{Simulation Scenario : Proximal Sensing of the Bunny}\label{subsec:proximal_bunny_sim}
In practice, robots must sense obstacles and avoid collisions online. However, training and generating the implicit surfaces offline is too slow; instead, we learn Gaussian CBF implicit surfaces directly from local data in real time.

\begin{figure}[!t]
\centering
 \subfloat{\includegraphics[width=0.5\linewidth]{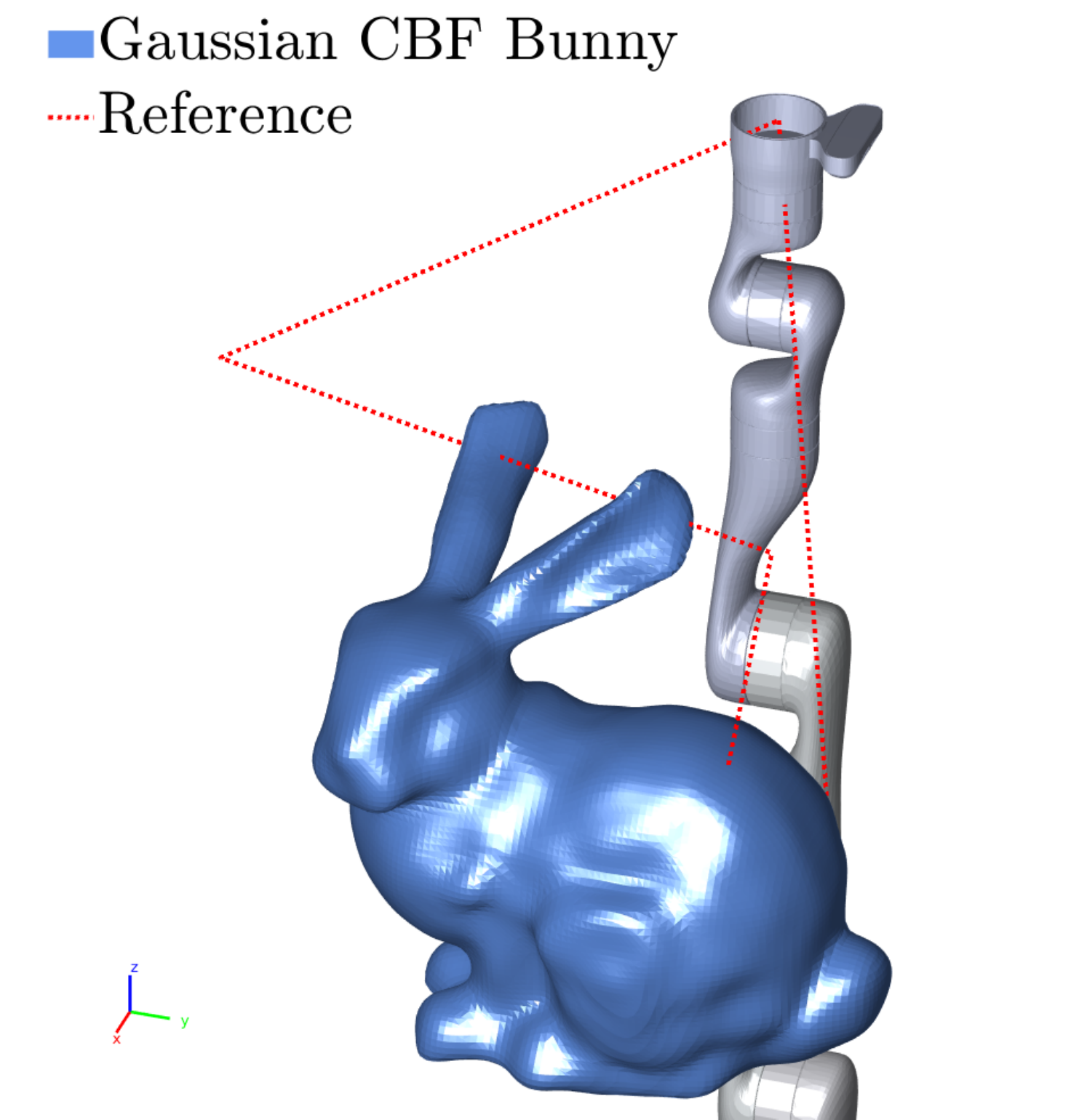}}
 \subfloat{\includegraphics[width=0.5\linewidth]{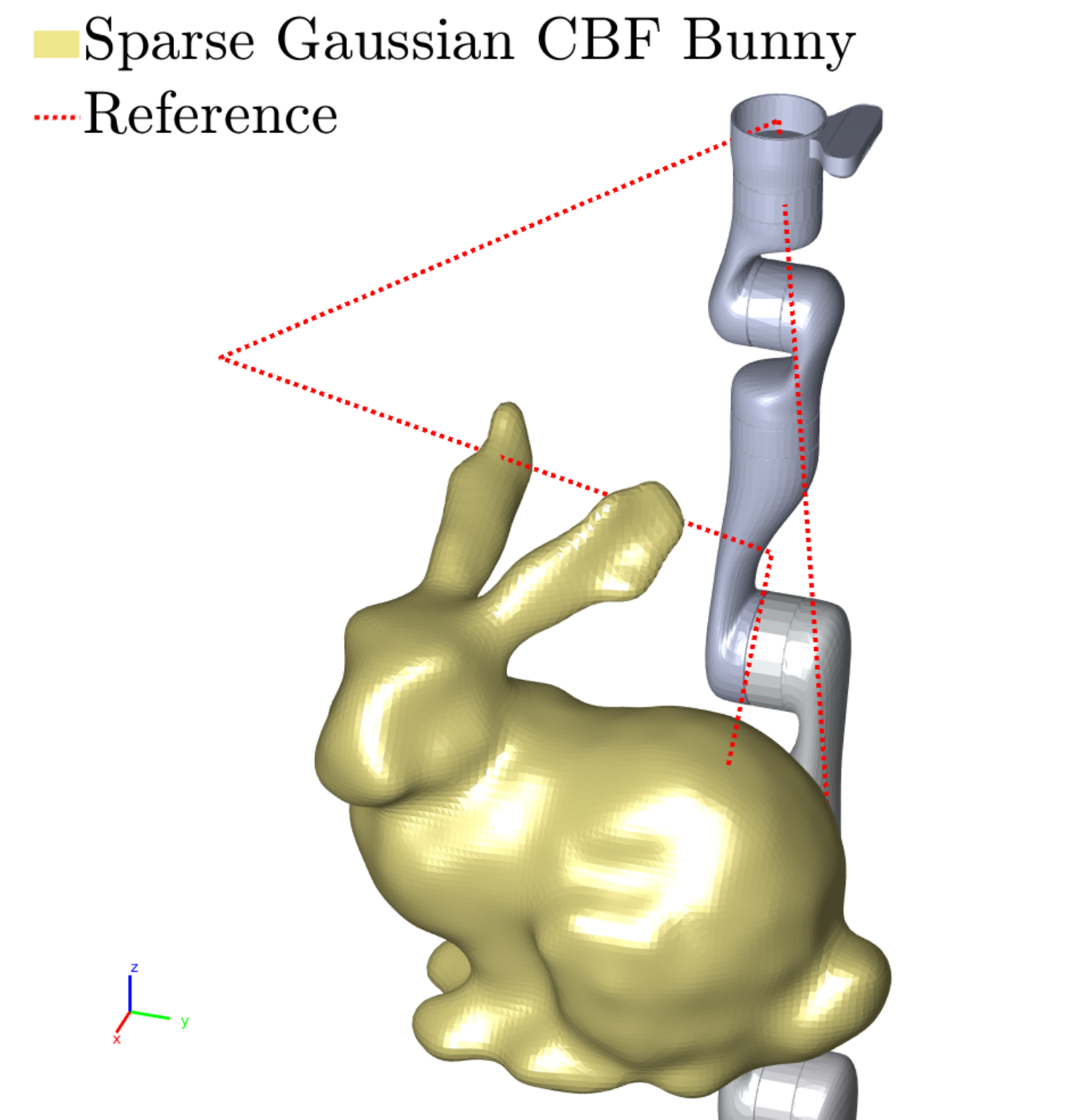}}
\caption{The reference trajectory passes through the bunny's back and ear regions for both synthesized CBFs without (left) and with (right) sparsity.}
\vspace{-0.3in}
\label{fig:bunny_robot_reference}
\end{figure}


\begin{assumption}\label{assump:sensor_model}
The proximal sensor can sense samples within its field-of-view (FOV) and scanning range, and provide the point cloud locations and surface normals.
\end{assumption}

\subsubsection{Proximal Sensing} We equip the manipulator’s end-effector with a sensor, modeled as a 3D LiDAR using a spherical cone configuration ($110^{\circ}$ FOV, $0.8\si{\m}$ range). 
A sample is detected if it is within the FOV and scanning range.

\subsubsection{Online Training \& Rectification} 
\begin{figure}[!t]
\centering
 \subfloat{\includegraphics[width=1\linewidth]{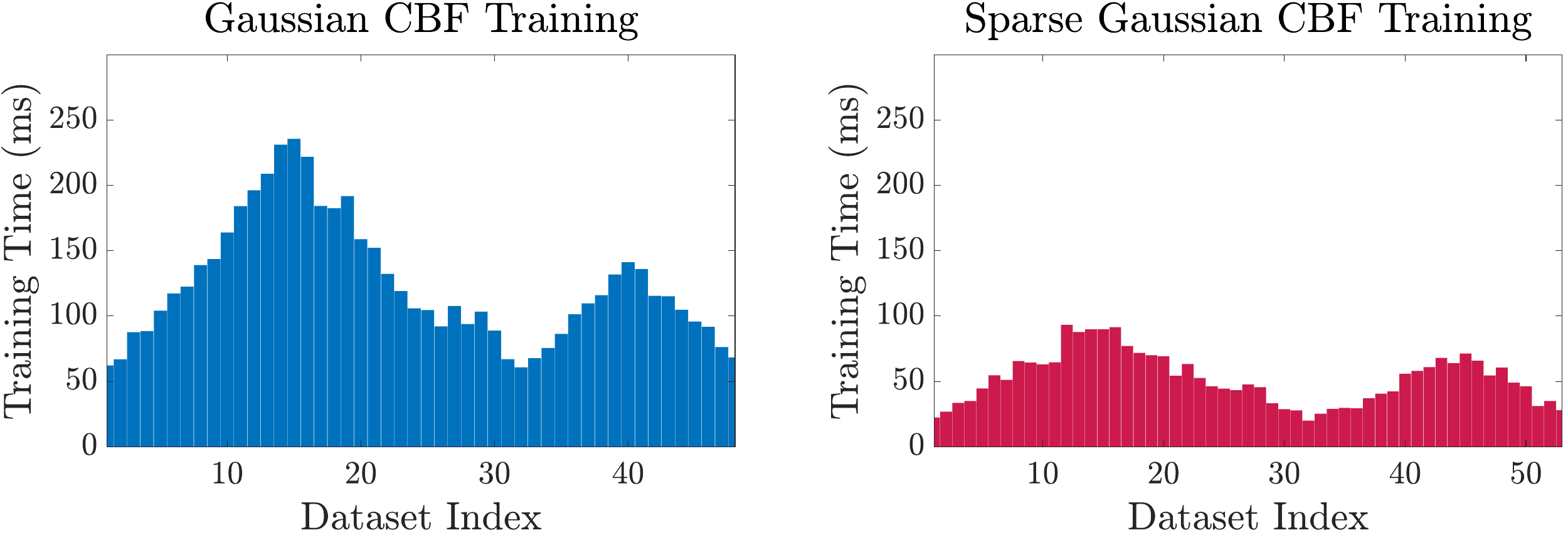}} \\
 \vspace{-.25cm}
 \subfloat{\includegraphics[width=1\linewidth]{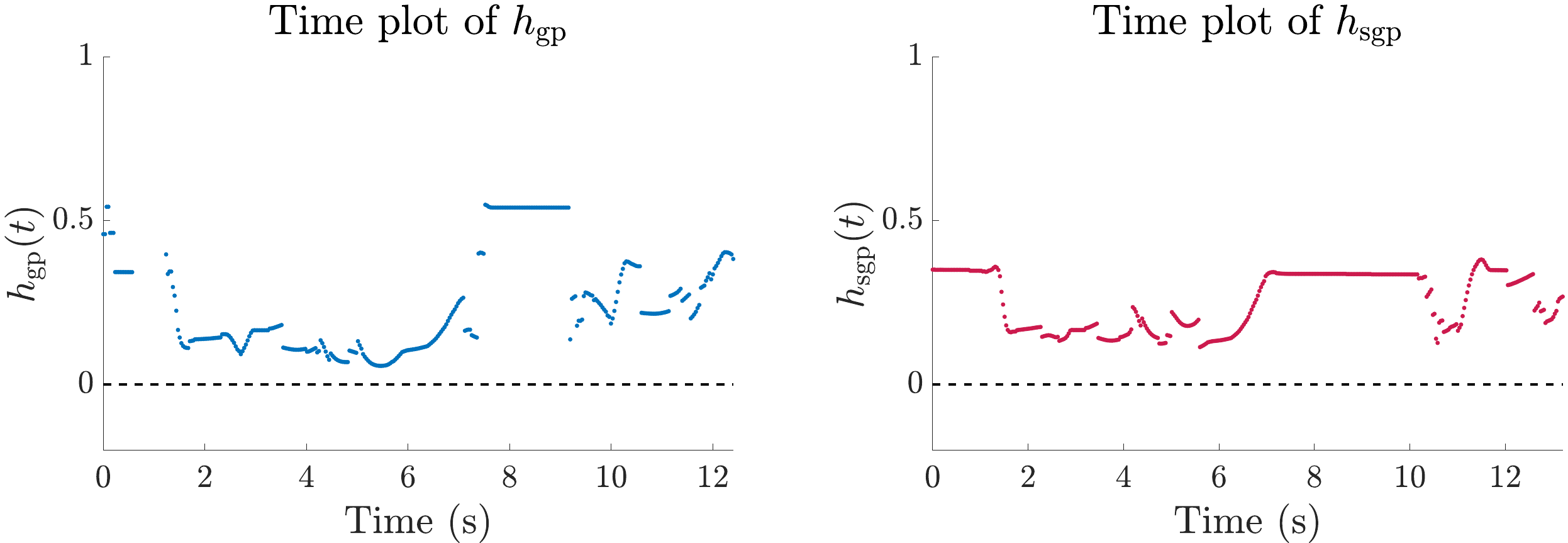}}
\caption{The training times per local dataset is shown for both the CBFs (top). The time plots (bottom) are non-negative, indicating no safety violations occurred.}
\vspace{-0.5cm}
\label{fig:bunny_local_training_time_h_array}
\end{figure}

The rectified control law is applied for both the Gaussian CBFs (Fig.~\ref{fig:bunny_robot_reference}). The manipulator follows the reference trajectory when safe, but relaxes tracking near the bunny to avoid collisions. Gaussian CBFs are trained online as samples arrive from local sensing: a local dataset is created if $\geq 100$ points are detected, where half the points are used as pseudo-inputs for sparse GP. The full GP formed $48$ datasets with an average training time of $124\si{\ms}$, while the sparse GP formed $53$ datasets with $52\si{\ms}$. 
Figure~\ref{fig:bunny_local_training_time_h_array} also shows the time plots of the CBFs, which are always non-negative, indicating no safety violations. The average per-query inference time at the end-effector is approximately $5\si{\ms}$ for both models.

\begin{figure}[!t]
\centering
 \subfloat{\includegraphics[width=0.5\linewidth]{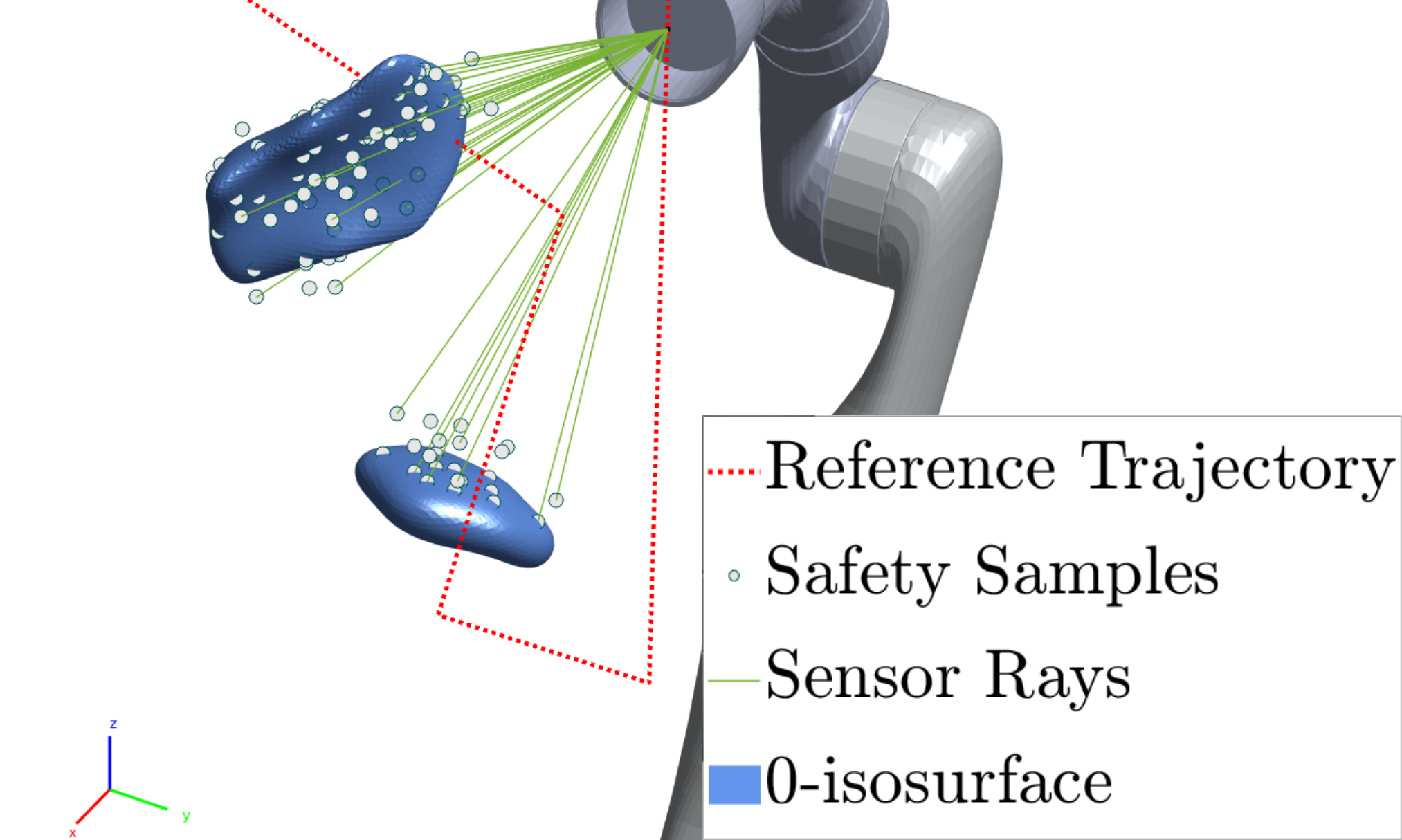}}
 \subfloat{\includegraphics[width=0.5\linewidth]{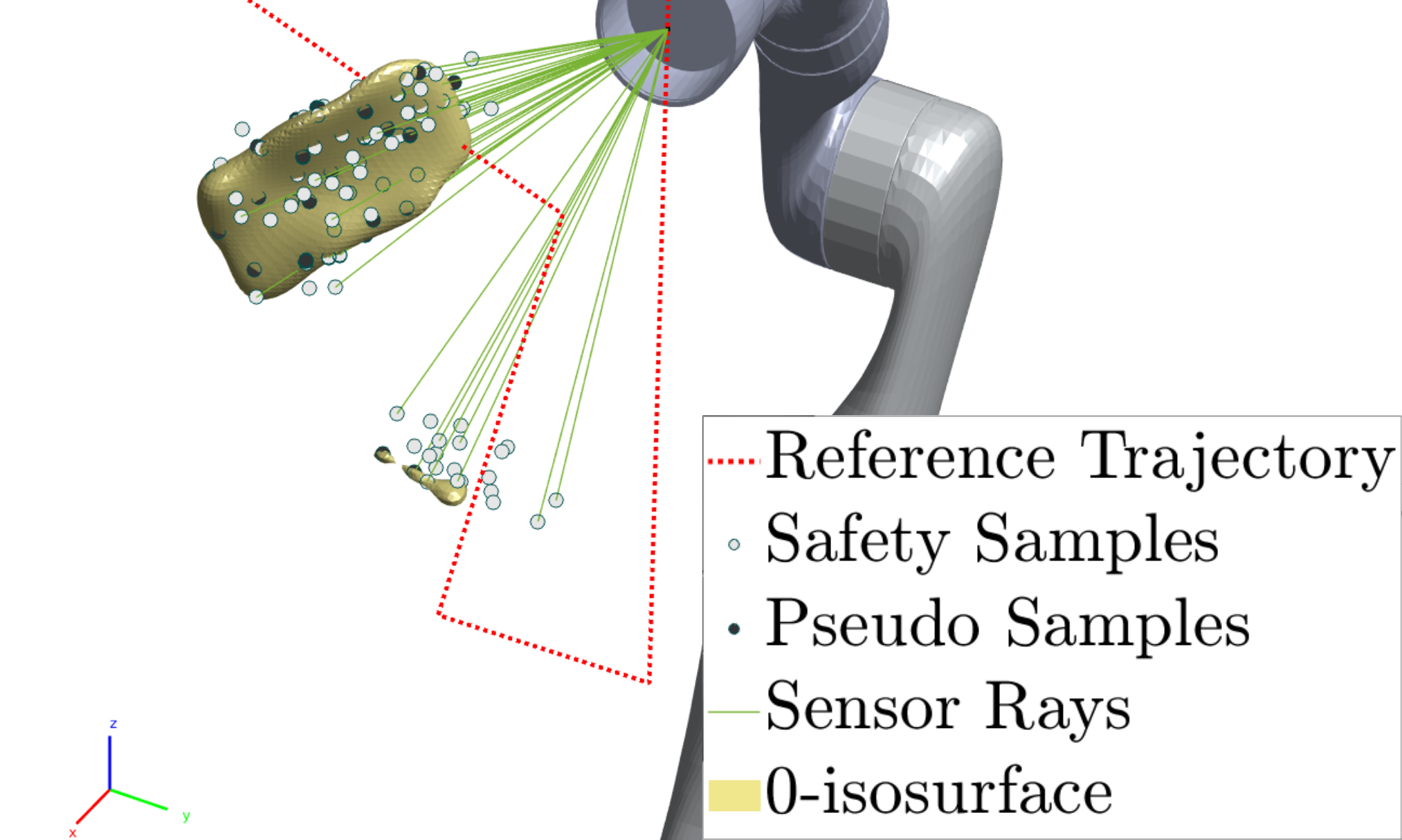}} \vspace{-0.3cm}\\
 \subfloat{\includegraphics[width=0.5\linewidth]{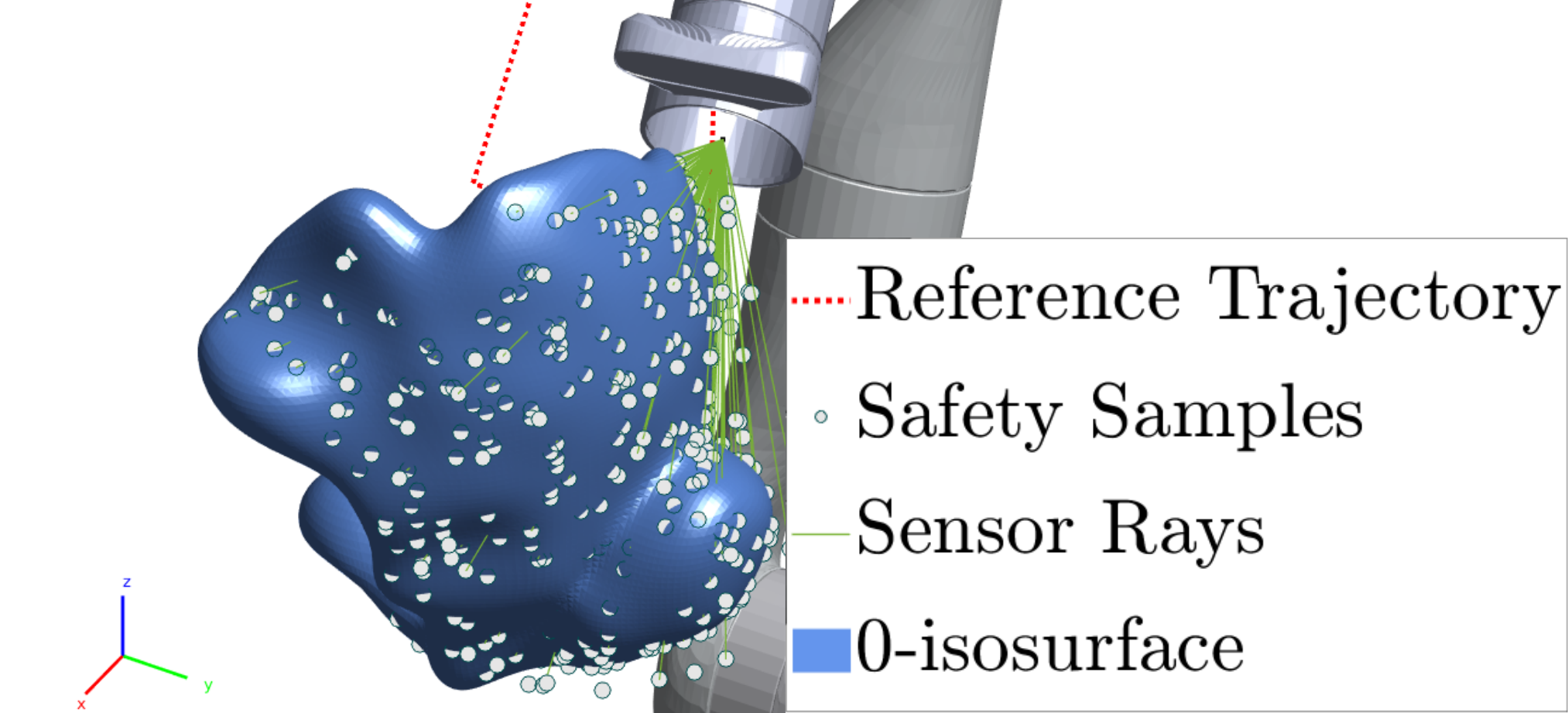}}
 \subfloat{\includegraphics[width=0.5\linewidth]{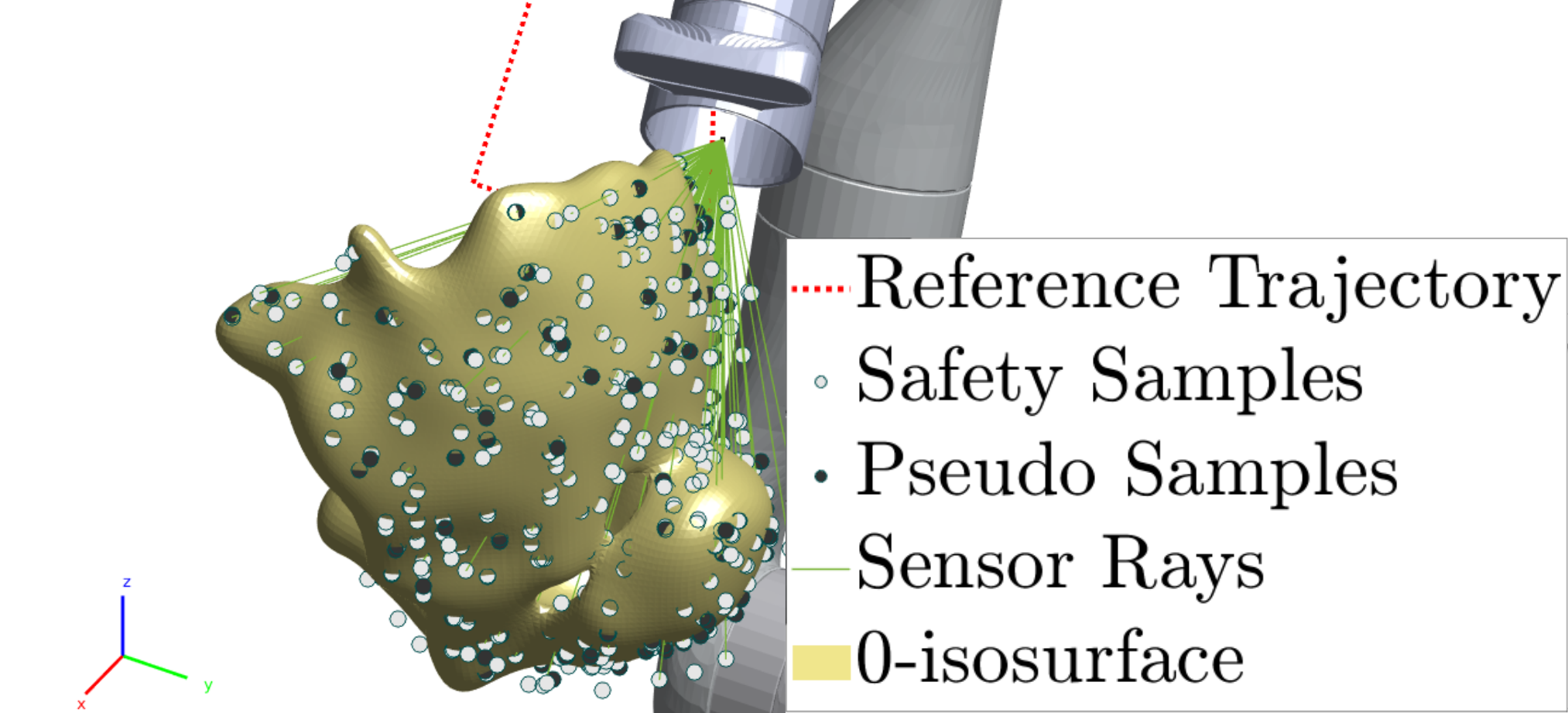}} \vspace{-0.3cm}\\
 \subfloat{\includegraphics[width=0.5\linewidth]{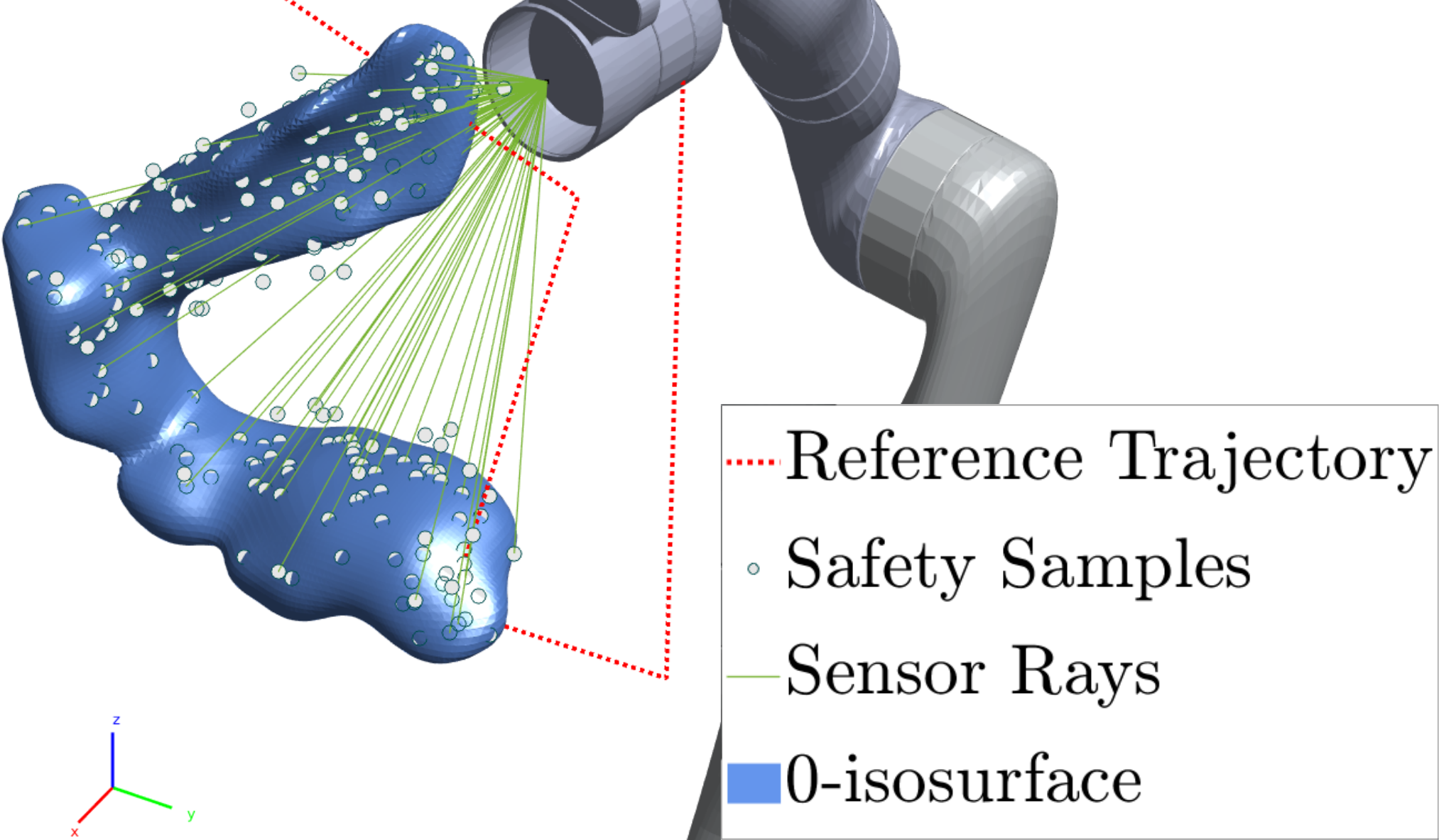}}
 \subfloat{\includegraphics[width=0.5\linewidth]{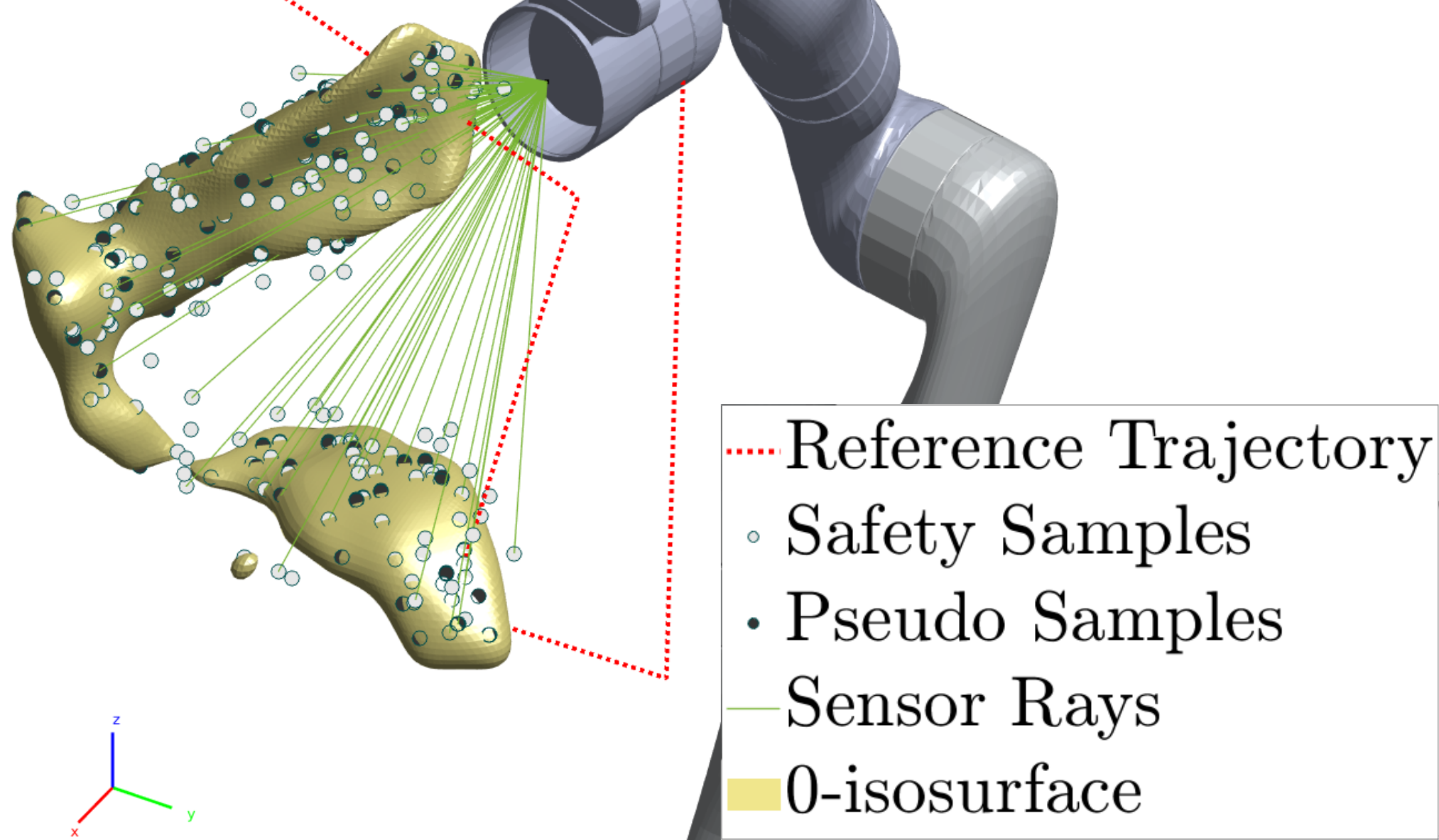}} \vspace{-0.3cm}\\
 \subfloat{\includegraphics[width=0.5\linewidth]{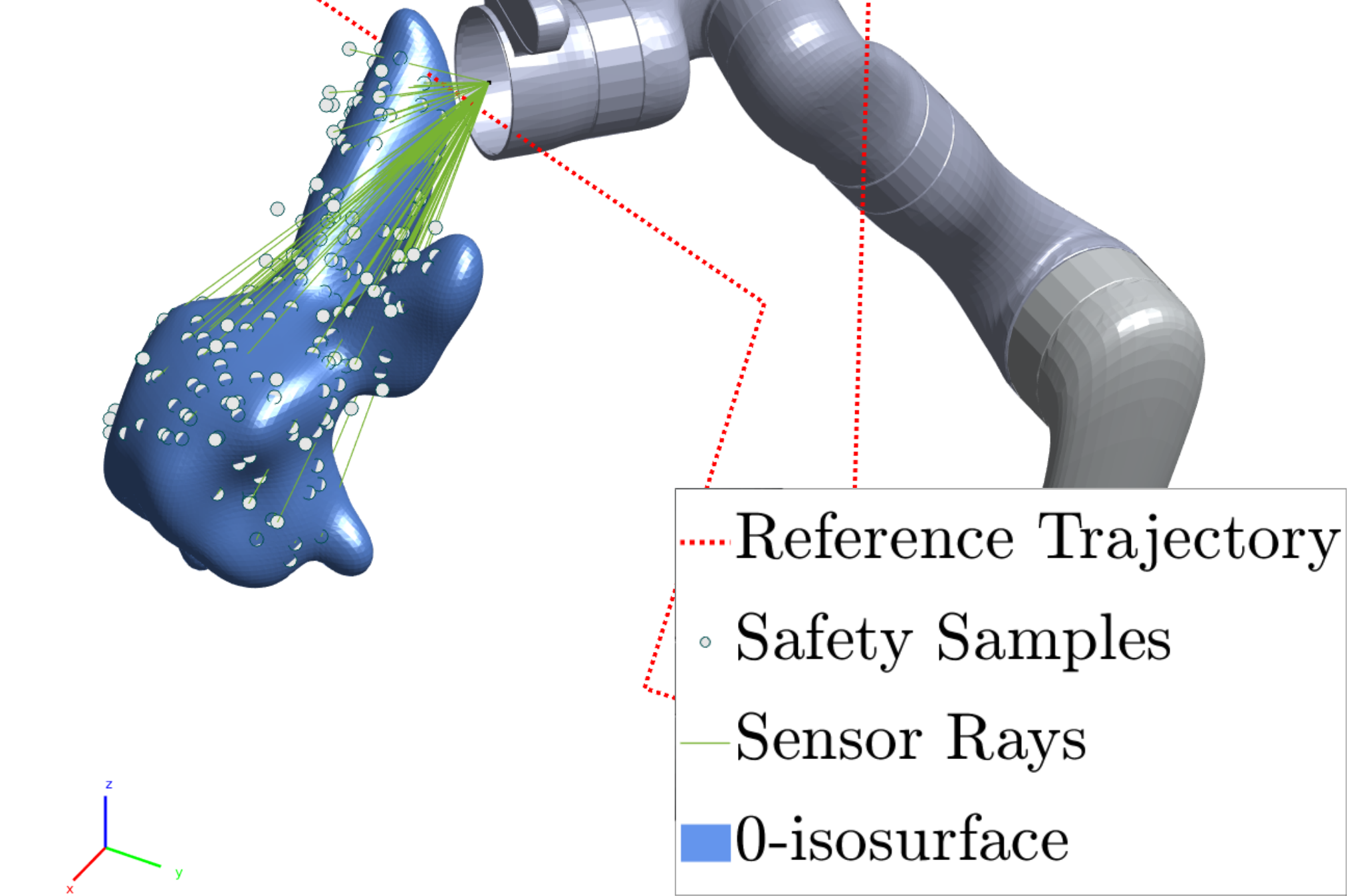}}
 \subfloat{\includegraphics[width=0.5\linewidth]{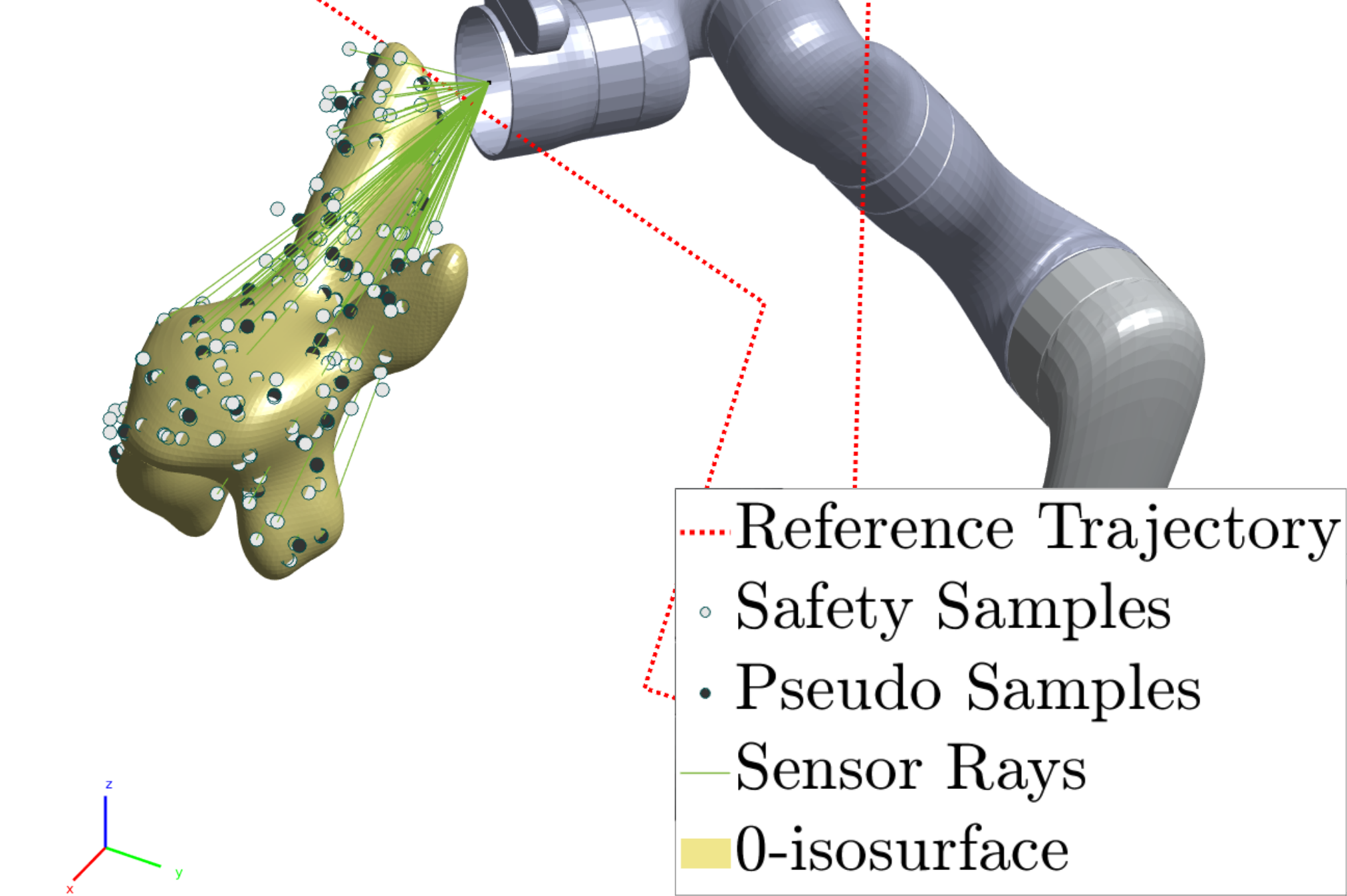}}
\caption{(Left) Gaussian CBFs trained online from local point cloud samples (white discs). (Right) Sparse Gaussian CBFs using pseudo-inputs (black discs) with local data; both show $0$-isosurfaces modeling the bunny surface.}
\vspace{-0.5cm}
\label{fig:gcbf_and_sgcbf_bunny_snapshots}
\end{figure}

\subsubsection{Discussion} Figure~\ref{fig:gcbf_and_sgcbf_bunny_snapshots} shows four simulation instances for both the synthesized CBFs. As expected, Gaussian CBF yields finer isosurfaces than the sparse variant. For example, Gaussian CBF captures the bunny’s neck and ear more precisely than the sparse Gaussian CBF. Yet the sparse model still delineates the safety boundary and prevents collisions. Notice that the data-driven generation of the CBFs is not limited to convex or connected surfaces. Disconnected isosurfaces can also be modeled using GPs.

\section{TEST CASE II : 3D QUADROTOR HARDWARE}\label{sec:test_case2_quad}
We next validate our method on a hardware quadrotor, a challenging platform due to its nonlinear, inherently unstable dynamics. We construct implicit surfaces as Gaussian CBFs from sensed data and apply them for safe control, ensuring collision avoidance around a chair.

\subsection{Experimental Hardware Overview}\label{subsec:experimental_hw_overview}
We use the Crazyflie 2.1, a $27\si{\gram}$ open-source quadrotor with 
a $15\si{\gram}$ payload limit \cite{crazyflie}. Due to its limited payload limit, it is unsuitable for any onboard depth sensing; thus, we generate Gaussian CBF surfaces offline. State estimation runs onboard aided with the help of an external infrared positioning system \cite{crazyflie}, which requires line-of-sight. For collision-avoidance experiments we use an IKEA ADDE chair, chosen for its large backrest gap with multiple holes, enabling state estimation even when flying beneath or behind it (otherwise the lighthouse signal is occluded). GP training (with and without sparsity) and control rectification 
are executed on the same ground station as discussed in \ref{subsec:gpis_bunny}.

\subsection{Chair Implicit Surface Modeling}\label{subsec:gpis_chair}
We first model the static chair for hardware experiments. As the Crazyflie 
cannot carry a depth or 3D LiDAR sensor, we use virtual point cloud data. The 
chair’s OBJ file is processed in \texttt{MeshLab} to 
extract $10{,}000$ point and surface normal samples. We train a GP with the 
Mat\'ern kernel (a generalization of the SE kernel) using $\nu = 3/2$:  
\begin{align}\label{eq:matern_kernel}
k_{\nu = 3/2}(\x_i,\x_j) = \sigma_f^2 
				\big( 1 + t  \big) 
				\exp \big(- t \big) + \delta_{ij} \sigma_\omega^2,
\end{align}
where $t = \frac{ \sqrt{3} \ \| \ \x_i - \x_j \| } {l} \in \R_{\geq 0}$, $\x_i, \x_j \in \Rn$, $i, j \in \{1, \dots, N\}$, $N$ is the number of samples, and $\sigma_f \in \R$, $l \in \R$, $\sigma_\omega$ are signal variance, length scale, and signal noise hyperparameters respectively. Unlike the SE kernel, which assumes infinite smoothness and may be unrealistic for physical phenomena, the Mat\'ern class allows explicit control of smoothness and better captures real-world structures \cite{math_interpolation_Stein1999}.

We downsample the chair's point cloud to $2{,}201$ points and use 
one-third as pseudo-inputs for the sparse GP. Training times 
were $4.8\si{\s}$ for GP and $2.72\si{\s}$ for the sparse GP, with 
$15$ iterations of optimization defined in
\eqref{eq:sparse_log_mll}.  

\begin{figure}[!b]
\centering
\includegraphics[width=1\linewidth]{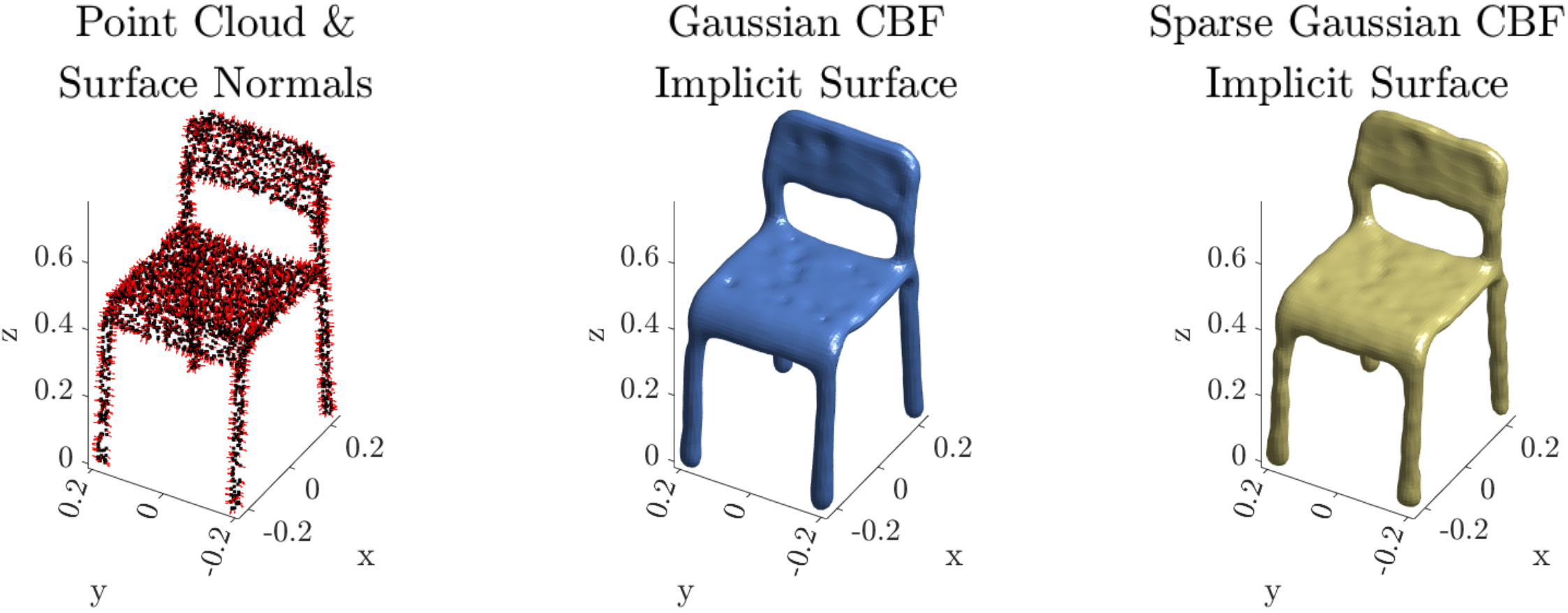}
\caption{The chair is modeled as the safe set boundary with Gaussian CBFs: (left) point cloud and surface normals for training, (middle) GP and (right) sparse GP implicit surface.}
\label{fig:gpis_chair}
\end{figure}

We use the same CBF candidate as \eqref{eq:candidate_gcbf_bunny}. Figure~\ref{fig:gpis_chair} shows the point cloud with surface normals and the GP-based implicit surfaces used as Gaussian CBFs. Gaps smaller than $0.05\si{\m}$ are ignored since the quadrotor cannot pass through them. The resulting Chamfer distances using \eqref{eq:chamfer_distance} between $\pchair$ and $\pcloudgp$/$\pcloudsgp$ are $0.078\si{\meter}$ and $0.0952\si{\meter}$, respectively. 

\subsection{Mat\'ern Kernel and Partial Derivatives}\label{subsec:matern_kernel_derivatives}
Here, we provide the equations of Mat\'ern kernel's Jacobian and Hessian for parameter $\nu = 3/2$. The first and second order partial derivatives of $t$ in \eqref{eq:matern_kernel} with respect to $\x$ at a query point $\xquery$ are,
\begin{align}
\dtdx \pipex 	\hspace{-0.3cm}&= 	\frac{ \sqrt{3} } {l} \frac{ \ \ (\xquery - \x)^{\top} }{ \| \xquery - \x \| }, \label{eq:t_jacobian}\\
\ddtddx \pipex 	\hspace{-0.3cm}&= 	\frac{ \sqrt{3} } {l} \frac{ \mathbf{I} }{ \| \xquery - \x\| } - \frac{ \sqrt{3} } {l} \frac{ (\xquery - \x)(\xquery - \x)^{\top} }{ \| \xquery - \x \|^3 } \label{eq:t_hessian}, 
\end{align}
where $\dtdx \in \R^{1 \times n}$, $\ddtddx \in \R^{n \times n}$, and $\mathbf{I} \in \R^{n \times n}$ is the identity matrix. The kernel's Jacobian and Hessian in \eqref{eq:matern_kernel} with respect to $\x$ at a query point $\xquery$ are,
\begin{align}
\dkdx \pipex \hspace{-0.35cm}&= - \sigma_f^2 \ t  \exp(-t) \dtdx \pipex, \label{eq:matern_jacobian}\\
\ddkddx \pipex  \hspace{-0.35cm}&= 	- \sigma_f^2 \ t  \exp(-t) \ddtddx \pipex \notag \\
				& \ \ \ + \sigma_f^2  ( t - 1 ) \exp(-t) \dtdx^{\top} \pipex \dtdx \pipex \hspace{-.2cm}, \label{eq:matern_hessian}
\end{align}
where \eqref{eq:t_jacobian} and \eqref{eq:t_hessian} are substituted.

\begin{figure*}[!t]
\centering
	\subfloat{\includegraphics[width=0.167\linewidth]{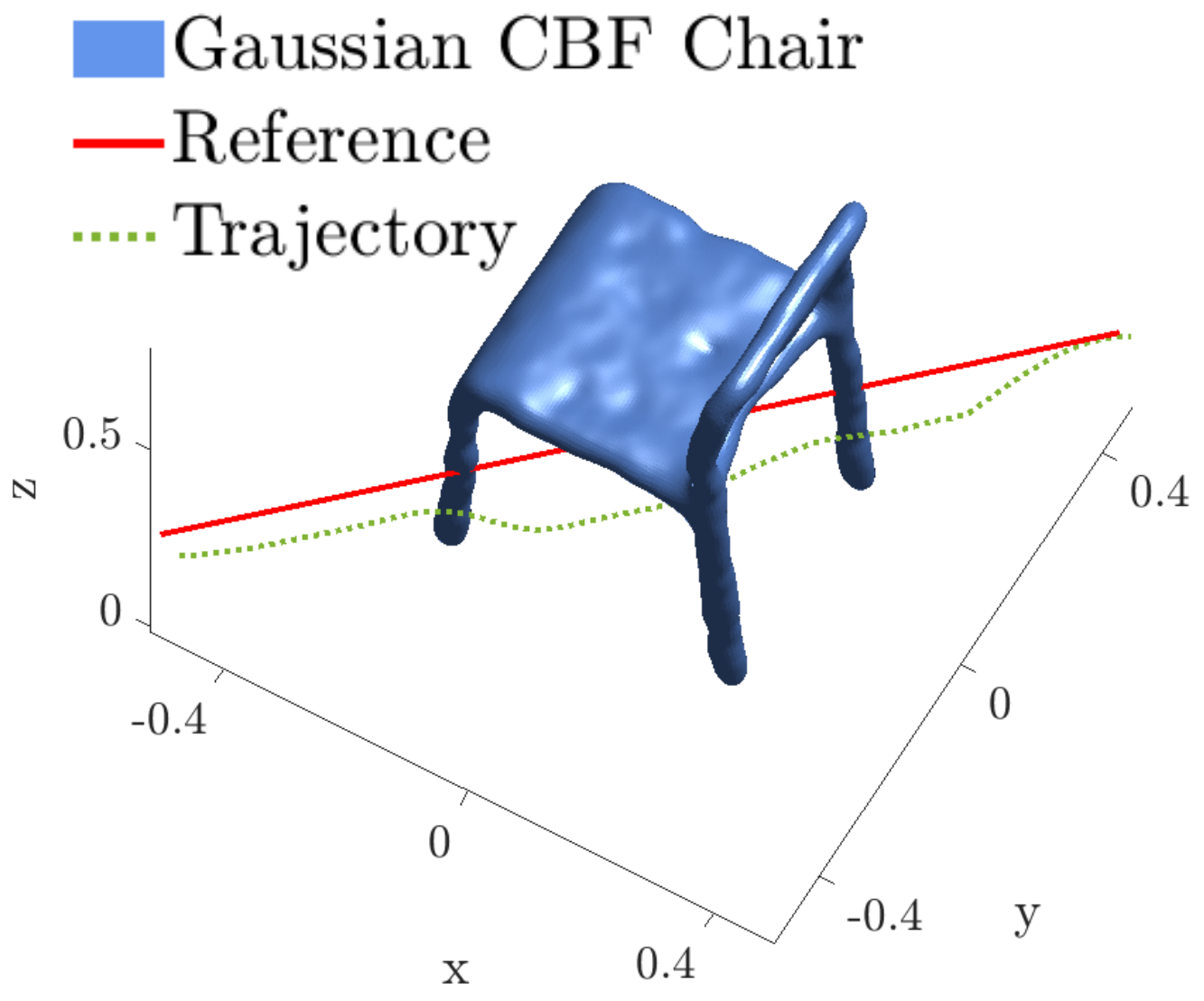}}
	\subfloat{\includegraphics[width=0.167\linewidth]{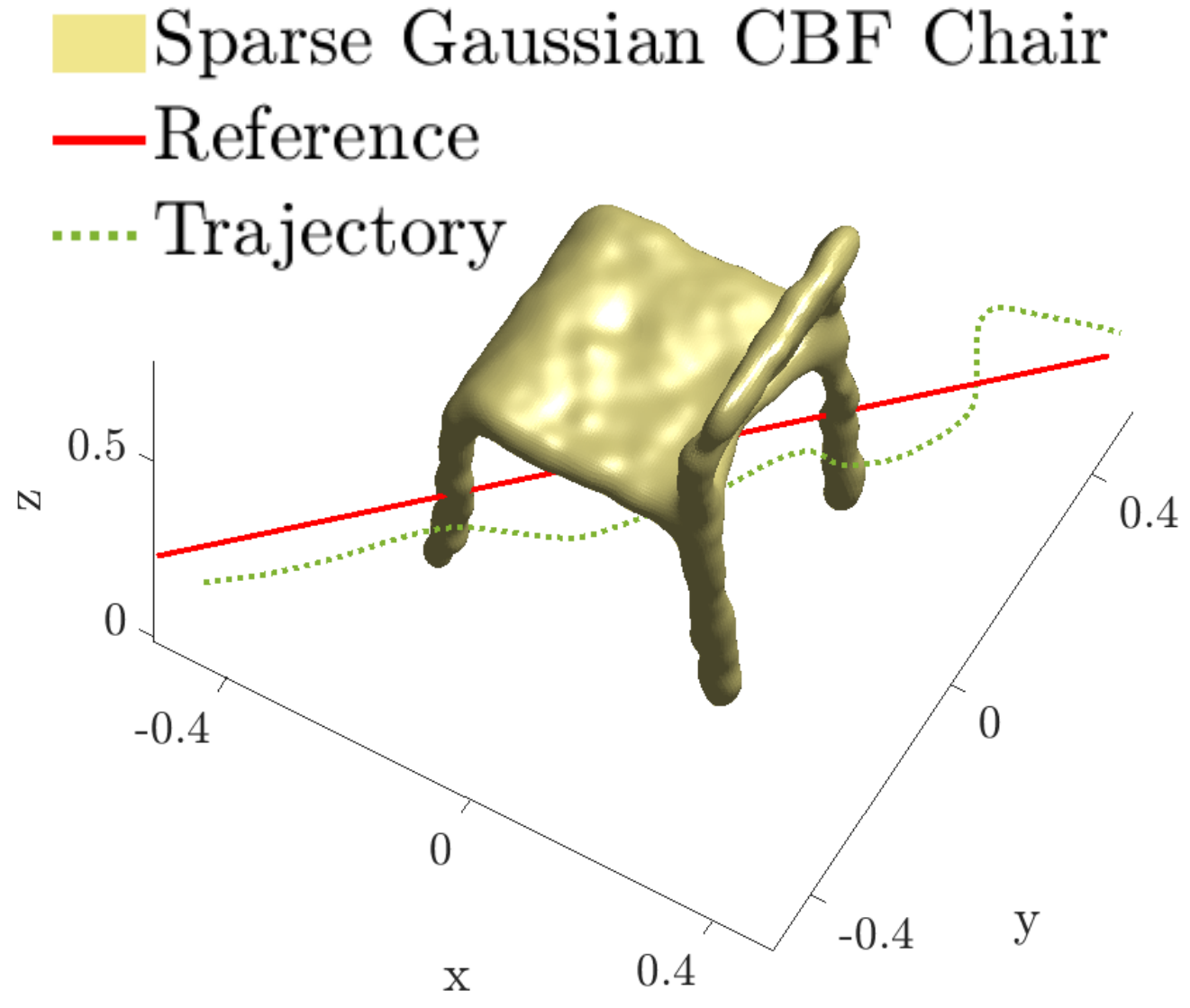}}
    \subfloat{\includegraphics[width=0.167\linewidth]{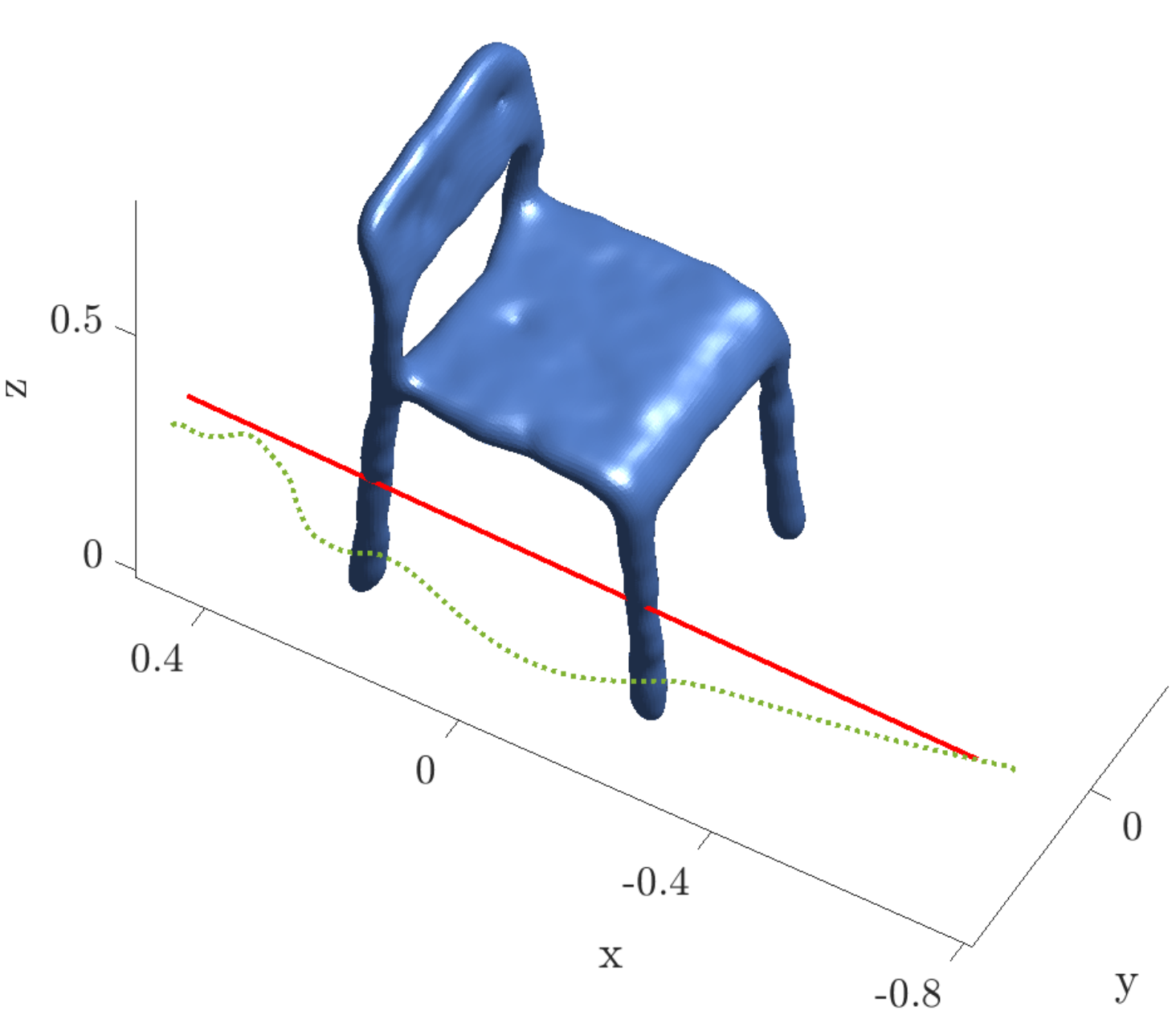}}
	\subfloat{\includegraphics[width=0.167\linewidth]{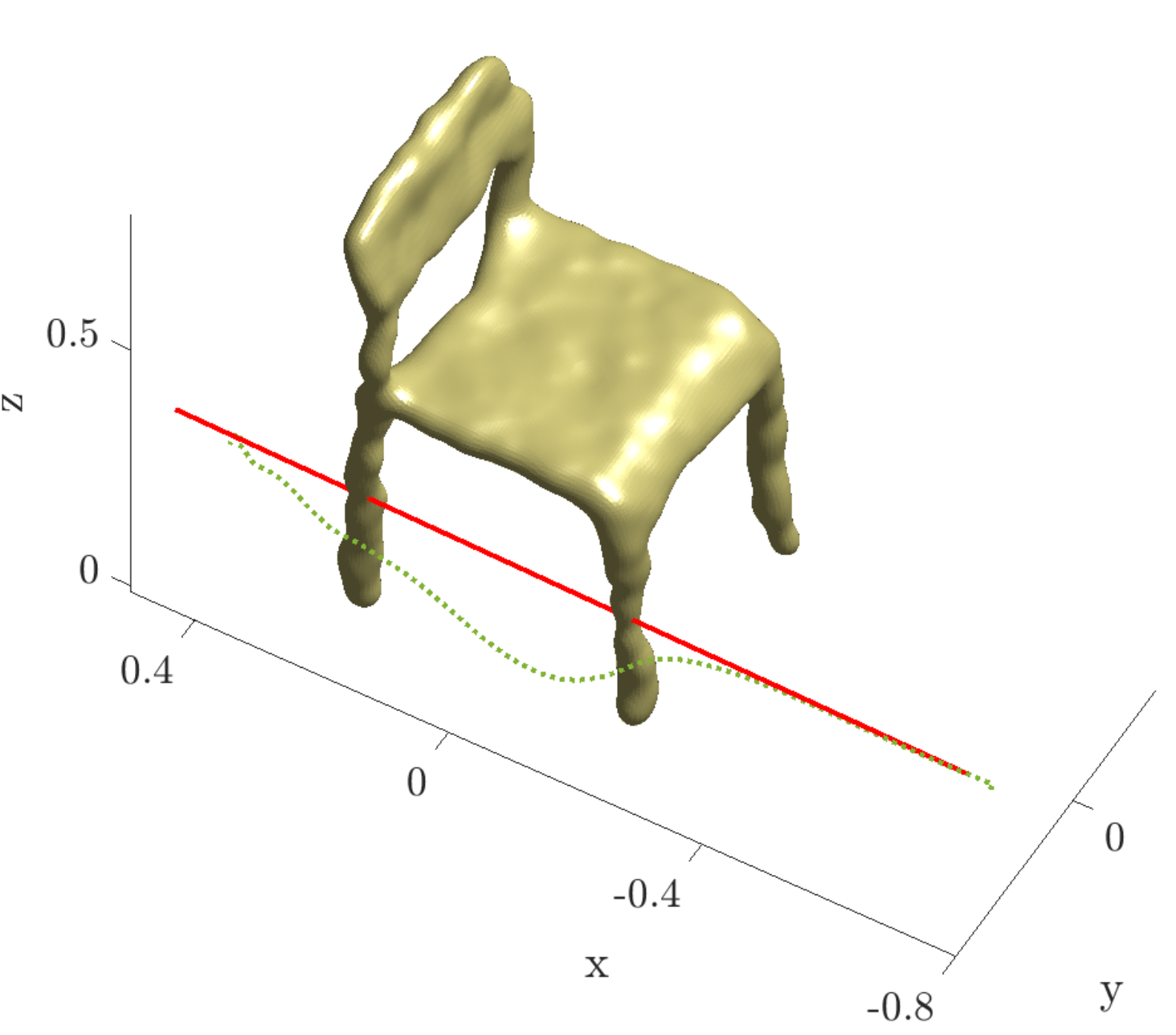}}
    \subfloat{\includegraphics[width=0.167\linewidth]{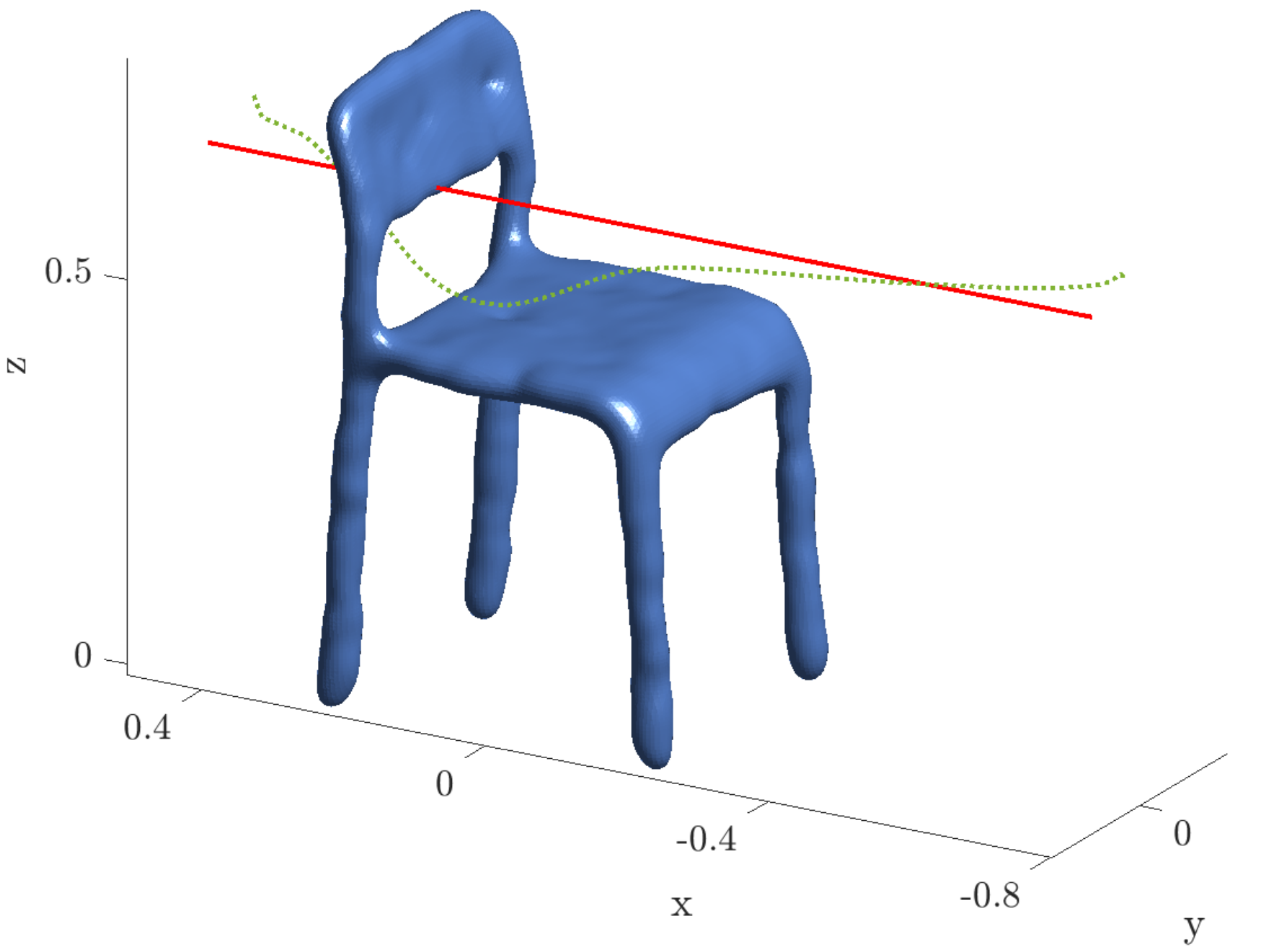}}
	\subfloat{\includegraphics[width=0.167\linewidth]{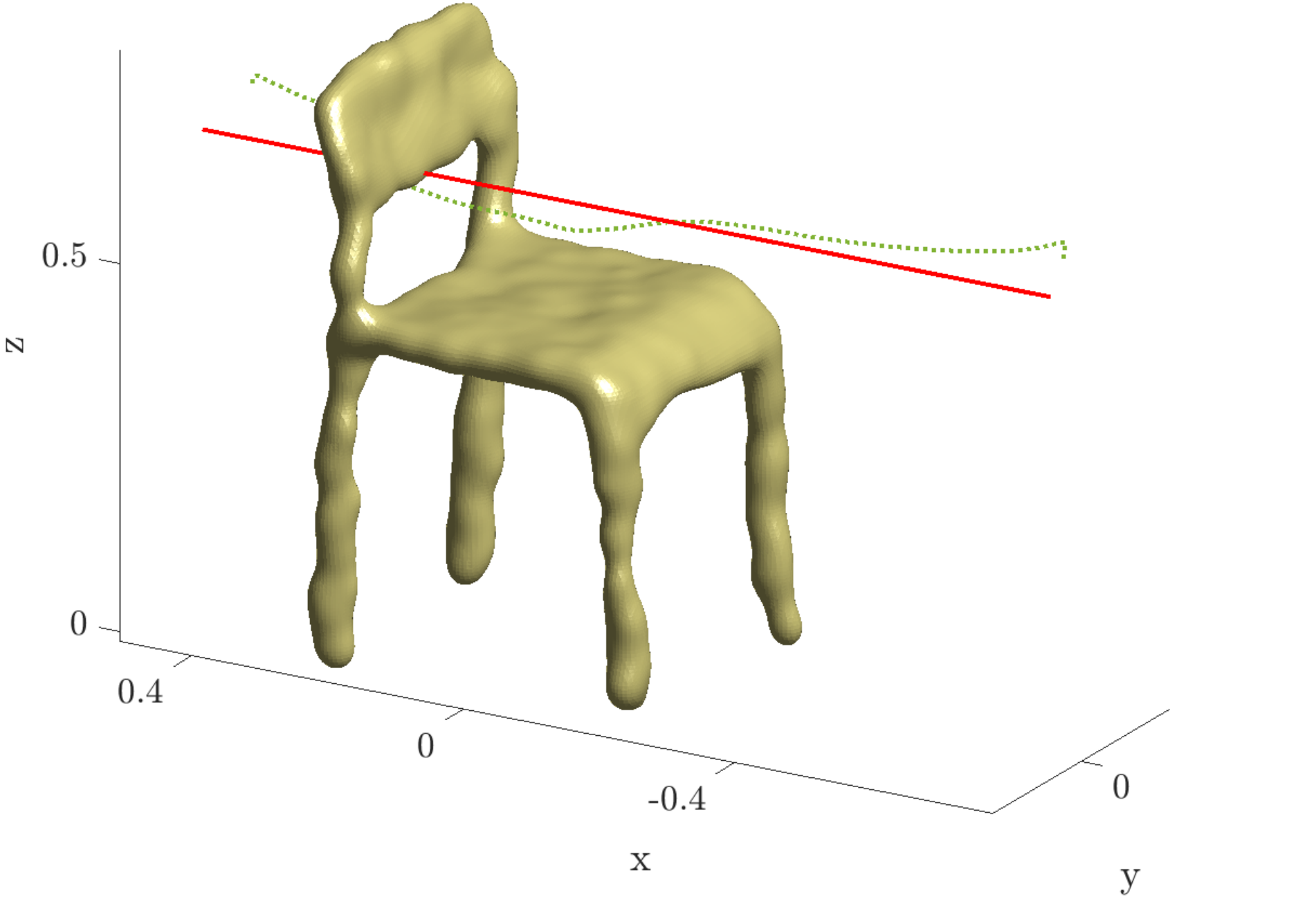}} \\
	\subfloat{\includegraphics[width=0.167\linewidth]{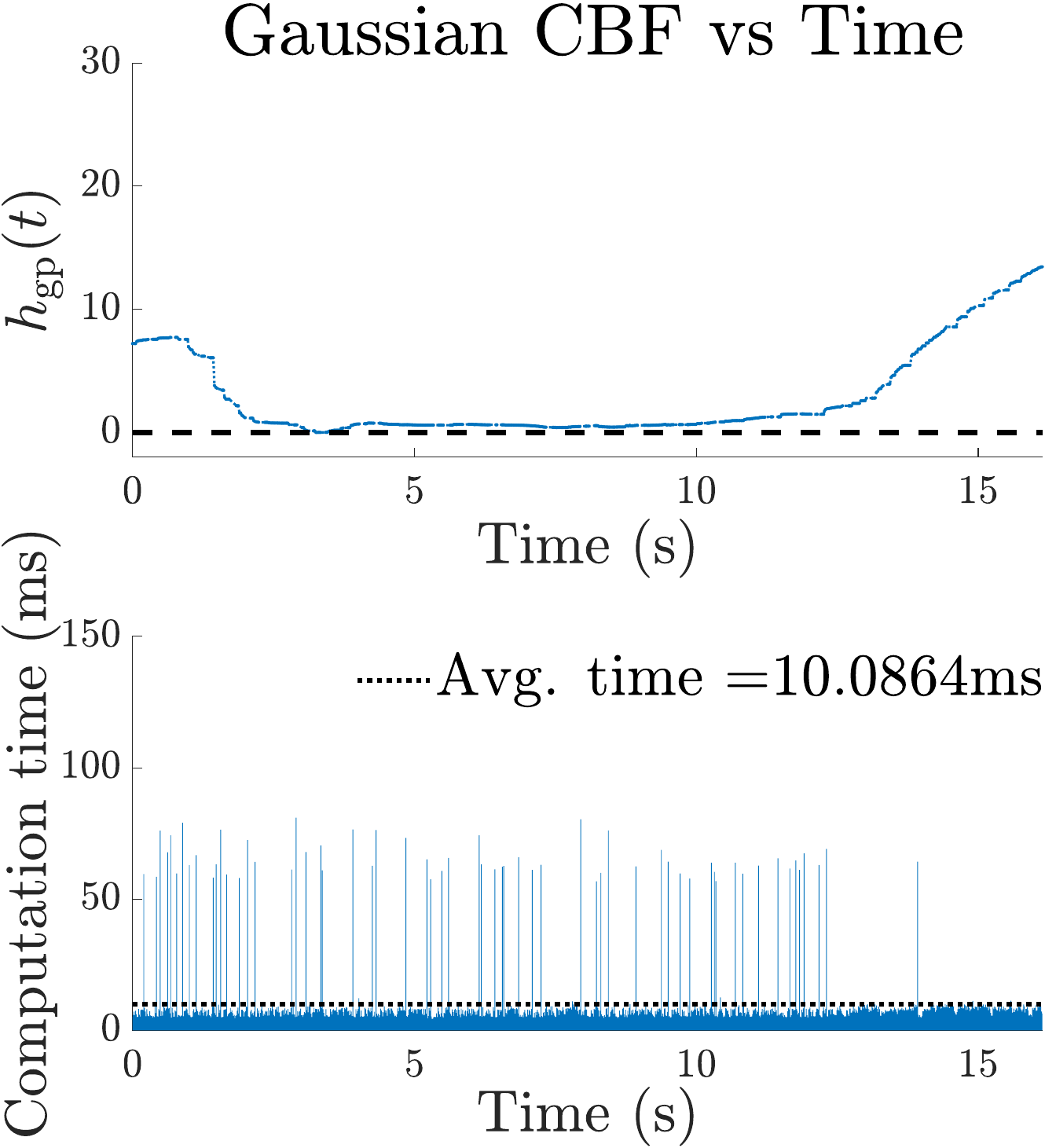}}
	\subfloat{\includegraphics[width=0.167\linewidth]{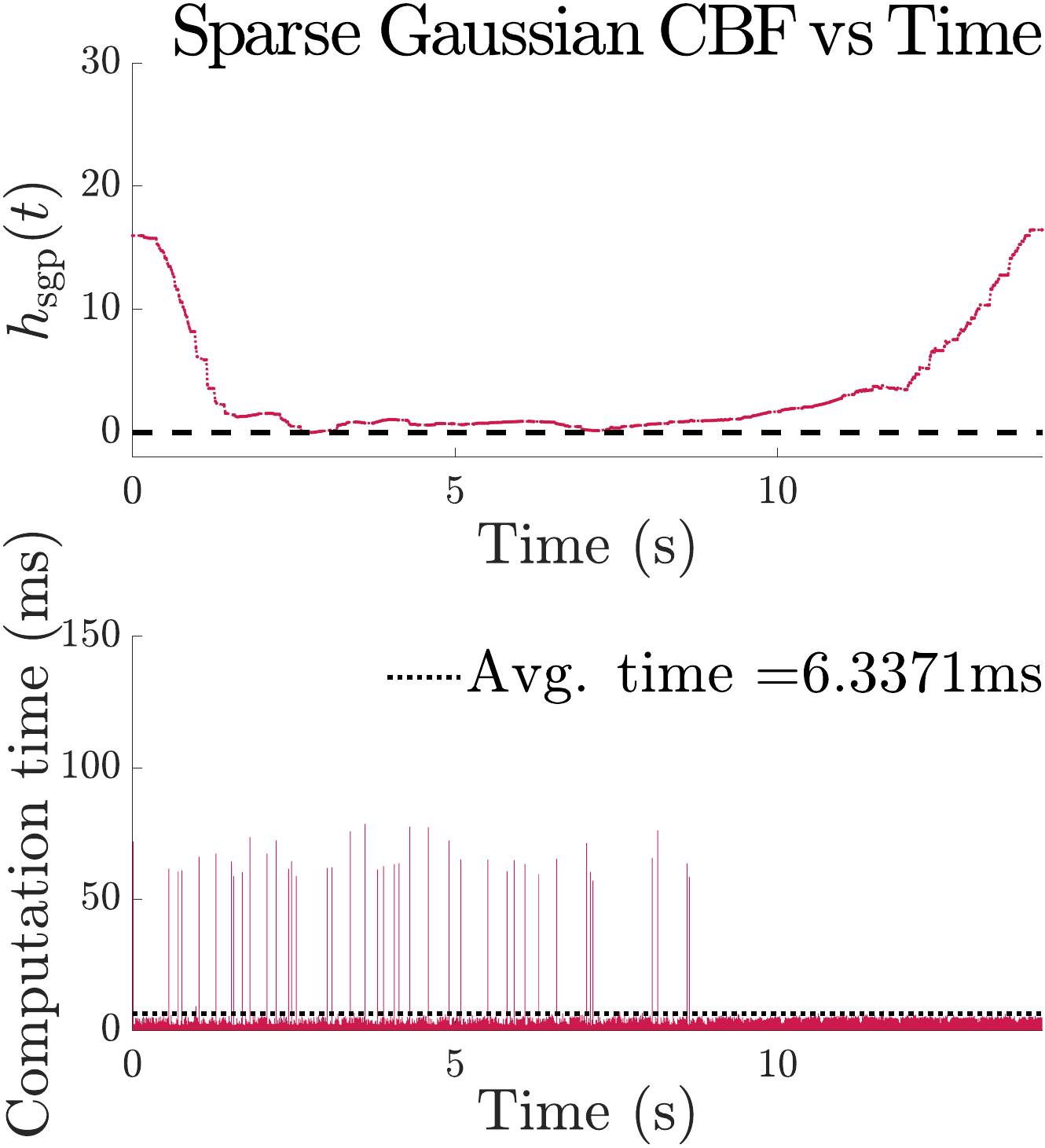}}
	\subfloat{\includegraphics[width=0.167\linewidth]{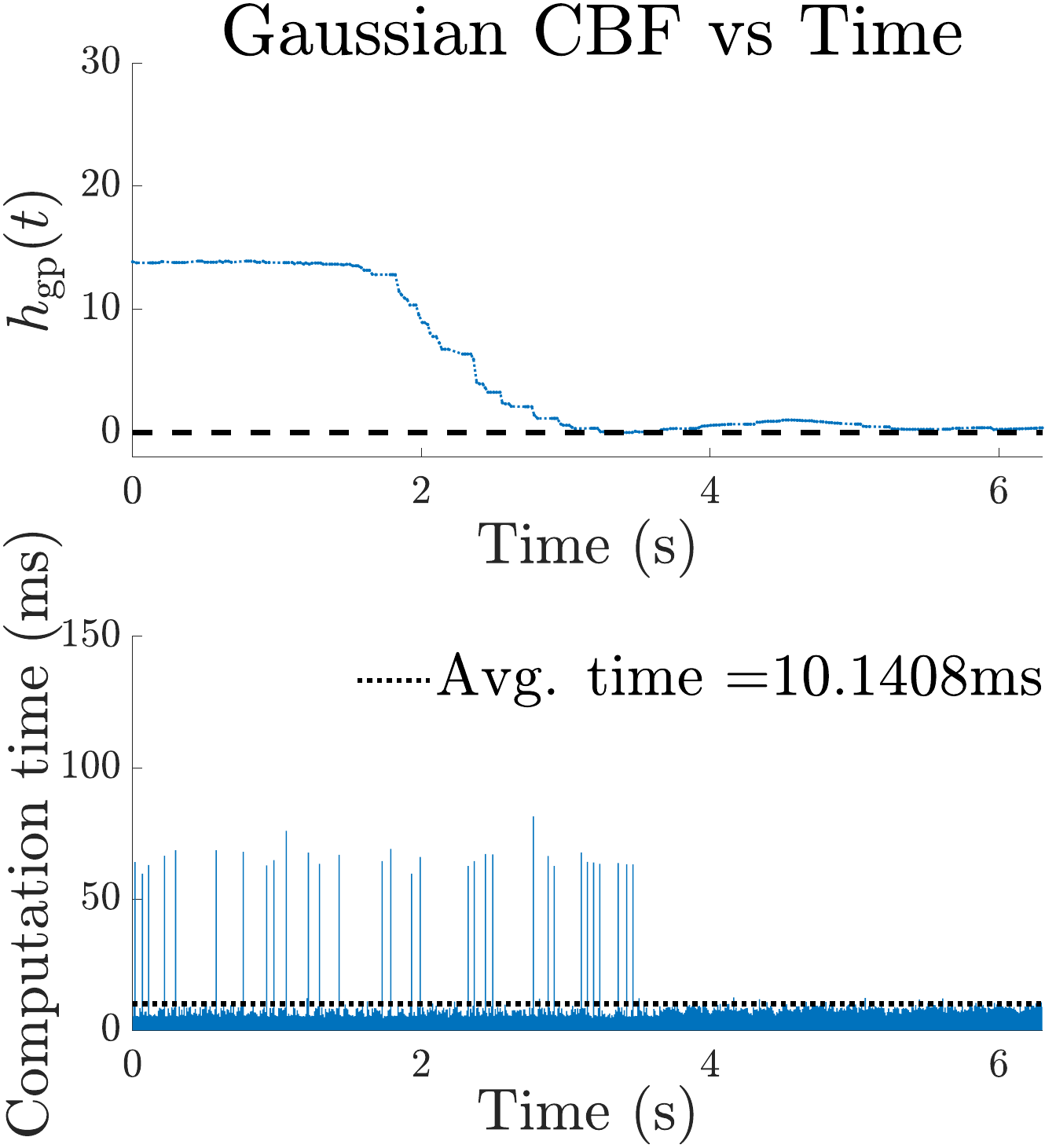}}
	\subfloat{\includegraphics[width=0.167\linewidth]{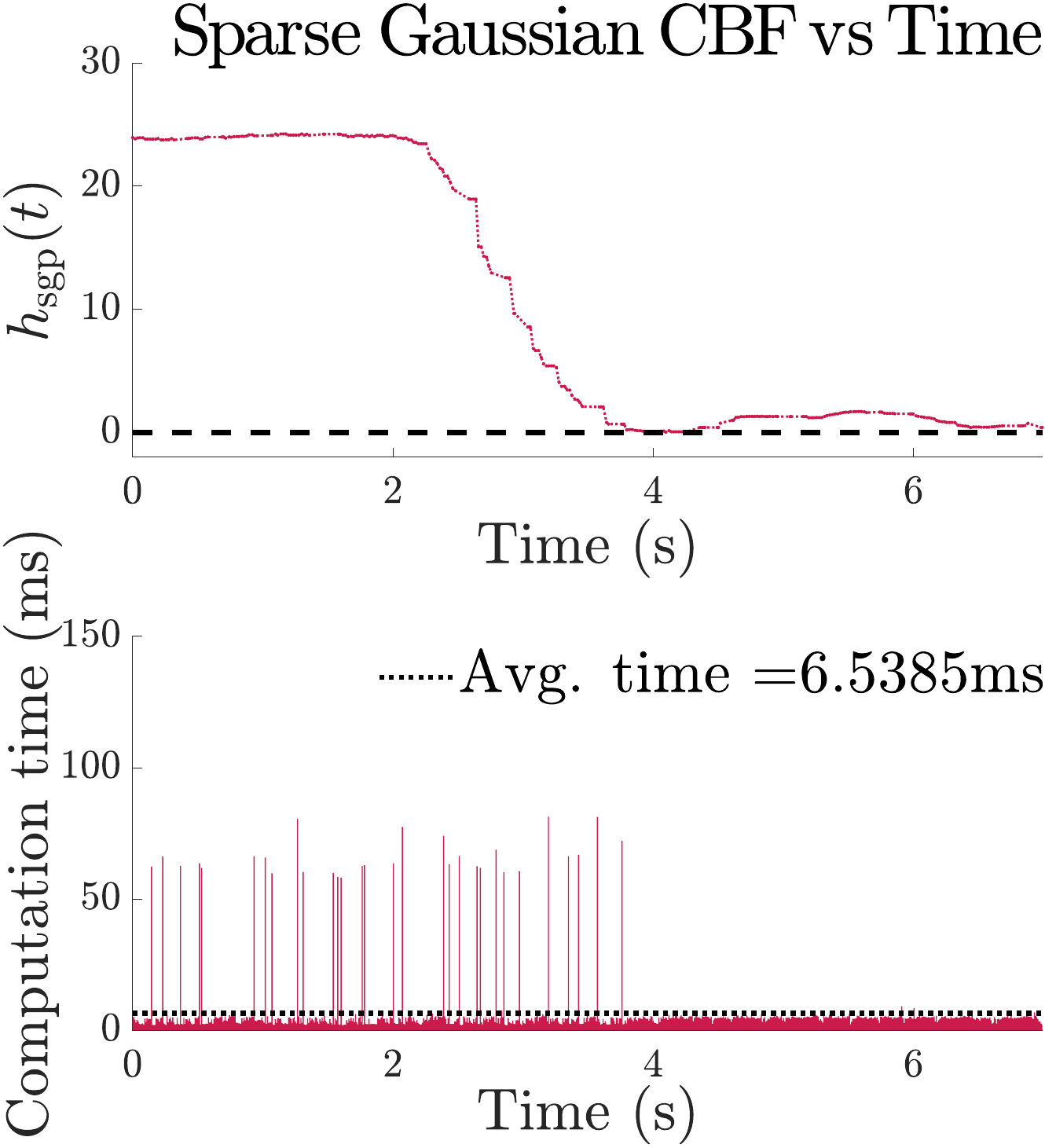}}
	\subfloat{\includegraphics[width=0.167\linewidth]{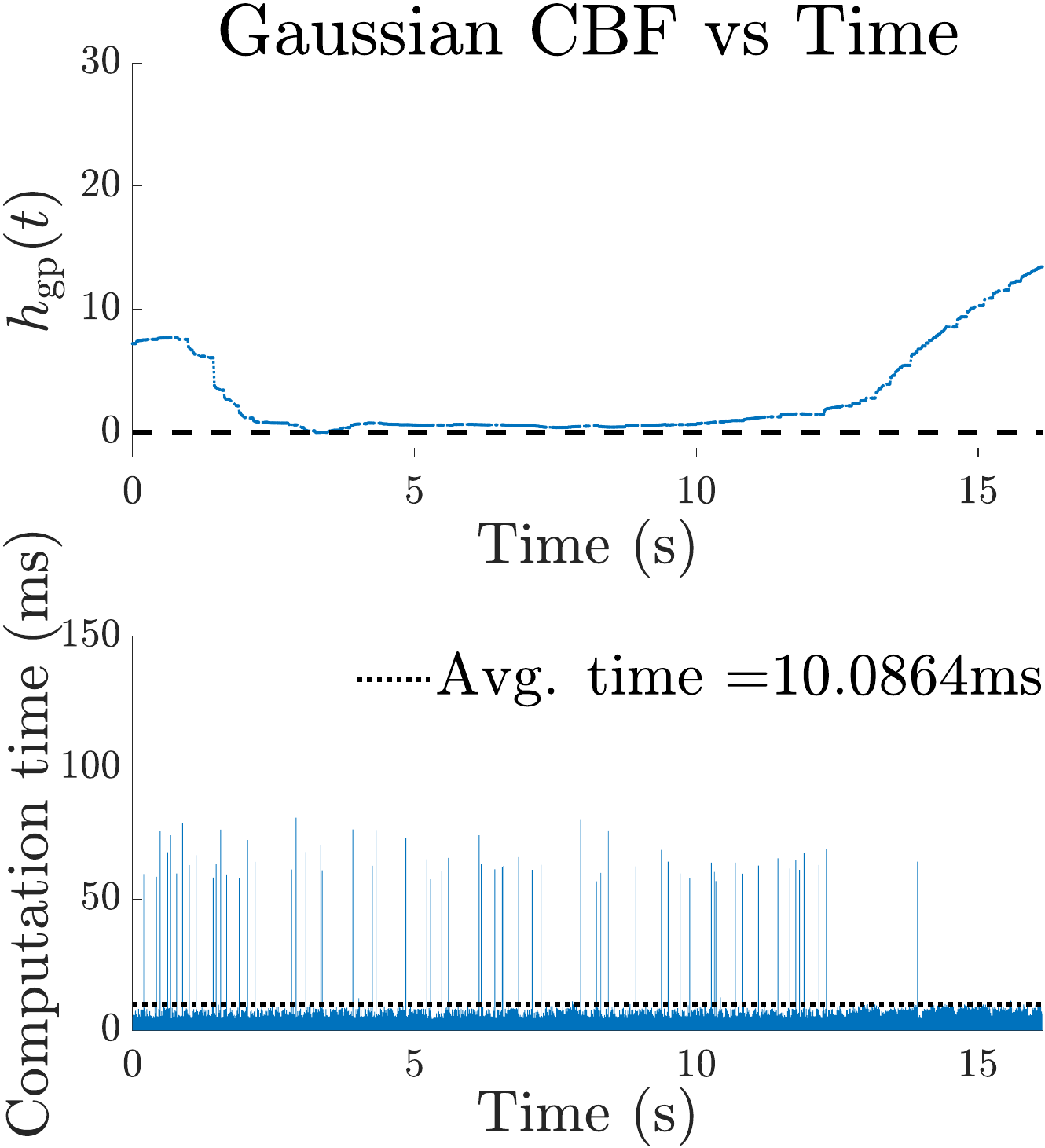}}
	\subfloat{\includegraphics[width=0.167\linewidth]{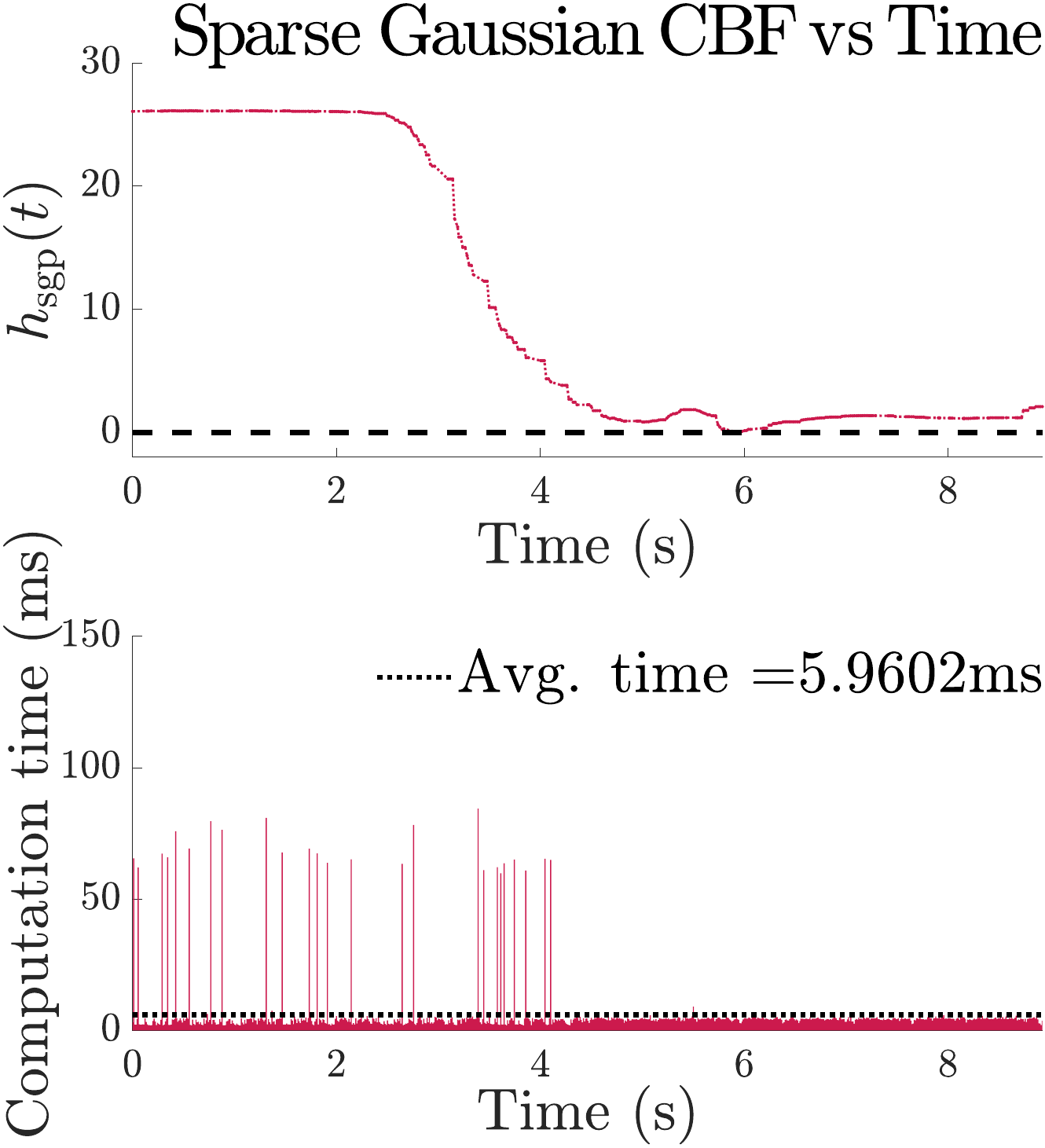}}
\caption{Gaussian (odd columns) and sparse Gaussian (even
columns) CBFs in three unsafe quadrotor scenarios: 
(i) diagonal through chair legs,
(ii) straight through front–rear legs, 
(iii) through the headrest. 
Each column shows the trajectories, CBF time plots, and compute times; in all cases safety was maintained, with average inference of $11\,\si{\ms}$ (GP) and $6.5\,\si{\ms}$ (sparse).}
\vspace{-0.5cm}
\label{fig:performance_gcbf_all}
\end{figure*}

\subsection{Safe Control Synthesis for Quadrotor}\label{subsec:safe_control_quadrotor}
We use thrust-attitude setpoints to control the Crazyflie \cite{ quadrotor_trajectory_generation_control_Mellinger2012}. We use the same second-order integrator and rectification process as  \cite{cbf_teleoperation_Xu2018}. The relative degree for the second-order integrator is $\rho = 2$. Hence, the associated Lie derivatives for the Gaussian CBF (with and without sparsity) in (\ref{eq:candidate_gcbf_bunny}) are,
\begin{align}\label{eq:lie_derivatives_2_order}
L_f \hcdot 
	&= 	\big( \nabla \mu(\x) - \nabla \sigma^2(\x) \big) ^{\top} f(\x), \notag \\
L_f^2\hcdot 
	&= 	\alpha^{\top} f(\x) + \beta^{\top} f(\x), \\
L_gL_f\hcdot 
	&= \alpha^{\top} g(\x) + \beta^{\top} g(\x), \notag
\end{align}
where $\nabla \mu(\x) = \dmudx^{\top}$ and $\nabla \sigma^{2}(\x) = \dvardx^{\top}$ 
are the gradients of GP mean and variance (with and without sparsity), $\nabla f(\x) = \tfrac{\partial f(\x)}{\partial \x}$ is the Jacobian of $f(\x)$, $\alpha = f(\x)^{\top} \big( \mathbf{H}_{\mu} (\x) - \mathbf{H}_{\sigma^2} (\x) \big)$, and $\beta = \big( \nabla \mu(\x) - \nabla \sigma^2(\x) \big) ^{\top} \cdot \nabla f(\x)$. The corresponding Hessians of both Gaussian and sparse Gaussian CBFs, $\mathbf{H}_{\mu}(\x)$ and 
$\mathbf{H}_{\sigma^2}(\x)$, are:
\begin{align*}
\mathbf{H}_{\mu_{\mathrm{gp}/\mathrm{sgp}}}(\x) &= \sum_i a_{\mathrm{gp}/\mathrm{sgp}}^{i} \ddkiddx, \\[-2ex] 
\mathbf{H}_{\sigma^2_{\mathrm{gp}/\mathrm{sgp}}}(\x) &= 
	- 2 \mathbf{A}_{\mathrm{gp}/\mathrm{sgp}} - 2 \sum_i b_{\mathrm{gp/\mathrm{sgp}}}^i(\x) \ddkiddx,
\end{align*}
where 
$a_{\mathrm{gp}}^i$ and $a_{\mathrm{sgp}}^i$ 
are the $i^{\mathrm{th}}$ entries of 
$\y_N^{\top} \KbarInv \hspace{-.2cm}\in \R^{1 \times N}$ 
and 
$\hspace{-.2cm}\y_N^{\top} \big(\Q_M^{-1} \ \K_{MN} \big( \Lm_N + \sigma_y^2 \I_N \big)^{-1}\big)^{\top} \hspace{-.2cm}\in\hspace{-.1cm} \R^{1 \times M}\hspace{-.3cm}$
, respectively.
$b_{\mathrm{gp}}^i(\x)$
and $b_{\mathrm{sgp}}^i(\x)$
are the $i^{\mathrm{th}}$ entries of 
$\mathbf{k}(\x)^{\top} \KbarInv \hspace{-0.2cm}\in \R^{1 \times N}$
and 
$\mathbf{k}_M(\x)^{\top} \big( \K_M^{ \ -1} - \Q_M^{ \ -1} \big) \in \R^{1 \times M}$
, respectively.
$\mathbf{A}_{\mathrm{gp}} = \nabla \mathbf{k}(\x) \mathbf{ \overline{K} \hspace{0.1cm} }^{-1} \nabla \mathbf{k}(\x)^{\hspace{-0.05cm}\top} \in \R^{N \times N}$,  
$\mathbf{A}_{\mathrm{sgp}} = \hspace{-0.1cm} \nabla \mathbf{k}_M(\x) \big( \K_M^{-1} \hspace{-0.15cm}-\hspace{-0.cm} \Q_M^{-1} \big) \hspace{-0.05cm}\nabla \mathbf{k}_M(\x)^{\top} \in \R^{M \times M}$, 
$\nabla \mathbf{k}(\x) = \dkdx^{\top} \in \R^{n \times N}$, 
$\nabla \mathbf{k}_M(\x) = \dkdx^{\top} \in \R^{n \times M}$, and $\ddkiddx$ is the partial derivative of (\ref{eq:kernel_deriv}) with respect to $\x$. Since the relative degree is $2$, we adopt
higher-order CBFs, in particular the exponential CBF  \cite{cbf_theory_Ames2019control, cbf_exponential_Nguyen2016}. Using the Lie derivatives in \eqref{eq:lie_derivatives_2_order}, we can design a QP of the form \eqref{eq:sgcbf-qp_degree1} to yield rectified control inputs that guarantee forward invariance of the safe set.

\subsection{Scenario: Safe Autonomous Navigation}\label{subsec:safe_autonomous_navigation}
In this scenario, the Crazyflie was commanded to follow unsafe reference
trajectories that passed through the chair. The goal was to track the reference
while ensuring safety by preventing collisions. We synthesized the CBFs across three separate runs, each using a different unsafe reference trajectory. The references were generated by specifying a final
desired position, with the thrust-attitude controller computing the desired
setpoint at every sampling step. Gaussian CBF rectification was applied at
$50\,\si{\hertz}$.

We evaluated three unsafe reference trajectories for the Crazyflie: (i) flying diagonally through the chair legs, (ii) flying straight through the front-left and rear-left legs, and (iii) flying directly through the chair’s headrest. These scenarios enforced reference paths that would naturally result in collisions, providing diverse and challenging conditions for assessing the effectiveness of Gaussian and sparse Gaussian CBFs in ensuring safe navigation.

Figure~\ref{fig:performance_gcbf_all} shows the flights, CBF temporal 
plots, and average per-iteration compute times. In all three runs, the 
quadrotor relaxed trajectory tracking when near the chair to maintain safety, avoiding collisions. The $\hgp$ and $\hsgp$ traces are nonnegative, and the average inference time was $11\si{\ms}$ for Gaussian CBF and about $6.5\si{\ms}$ for the sparse variant.

\section{CONCLUSION}\label{sec:conclusion}
In summary, we used Gaussian CBFs to construct safe implicit surfaces, where the surface of a volumetric object is defined as the boundary of the safe set. 
We presented sparse Gaussian CBFs to take advantage of reduced computational complexity from $\bigOgp$ to $\bigOsgp$ ($N$ is number of training points, $M$ is number of pseudo-inputs). 
We presented two robotic test cases, firstly, in simulation for a $7$-DOF manipulator, and secondly, for a hardware quadrotor. 
For the first study, we estimated the Stanford bunny’s safety boundary as Gaussian CBFs (with and without sparsity), where the safe implicit surface was the bunny’s surface. 
We demonstrated safe proximal sensing without collision on the manipulator in simulation. 
For the quadrotor, we demonstrated safe 3D navigation on the Crazyflie hardware around a physical IKEA ADDE chair, using its OBJ model to extract point cloud and surface normal data. 
Even when reference trajectories were unsafe and would have intersected the chair, the Gaussian CBF ensured collision-free autonomous flight in all trials.

\bibliographystyle{ieeetr}
\bibliography{root_icra26}

\end{document}